\documentclass[10pt,twocolumn,letterpaper]{article}
\usepackage[pagenumbers]{cvpr}
\ifdefined\pdfminorversion
\else
  \AtBeginDvi{}
\fi
\usepackage{graphicx}
\usepackage{amsmath,amssymb}
\usepackage{booktabs,multirow,tabularx,array}
\usepackage{microtype}
\usepackage{xspace}
\usepackage{enumitem}
\usepackage{pifont}
\usepackage{stix} 
\newcommand{\benchmark}{EmbodiSafety\xspace}

\newcommand{\mode}[1]{\textsc{\MakeLowercase{#1}}}
\newcommand{\na}{--}

\newcolumntype{Y}{>{\raggedright\arraybackslash}X}
\graphicspath{{figures/}}

\usepackage{colortbl}
\definecolor{paperpanel}{RGB}{233,239,243}
\definecolor{paperscore}{RGB}{197,222,212}

\makeatletter
\patchcmd{\@maketitle}{\null}{\null\vspace{-5em}}{}{}
\makeatother
\definecolor{cvprblue}{rgb}{0.21,0.49,0.74}
\usepackage[pagebackref,breaklinks,colorlinks,allcolors=cvprblue]{hyperref}
\def\confName{CVPR}
\def\confYear{2027}
\title{EmbodiSafety: Benchmarking Risk Understanding\\and Response in Embodied AI}
\author{%
\resizebox{0.96\textwidth}{!}{Erfei Cui$^{2,*}$, Shengjia Zhao$^{4,*}$, Yan Qin$^{3,*}$, Kaixuan Wang$^{1,*}$, Junming Chen$^{1}$, Ping Luo$^{1,\dagger}$, Tianxing Chen$^{1,\dagger}$}\\[5pt]
{\small $^{1}$HKU \quad $^{2}$SJTU \quad $^{3}$HKUST (GZ) \quad $^{4}$SZU \qquad $^{*}$Equal contribution \quad $^{\dagger}$Corresponding authors}\\[5pt]
{\normalsize\textbf{Project page:} \href{http://embodisafety.github.io/}{\texttt{http://embodisafety.github.io/}}}
}
\hypersetup{
    pdftitle={EmbodiSafety: Benchmarking Risk Understanding and Response in Embodied AI},
    pdfauthor={Erfei Cui, Shengjia Zhao, Yan Qin, Kaixuan Wang, Junming Chen, Ping Luo, Tianxing Chen}
}
\begin{document}
\makeatletter
\g@addto@macro\@maketitle{%
\begin{minipage}{\textwidth}
\centering
\captionsetup{hypcap=false}
\includegraphics[width=\textwidth]{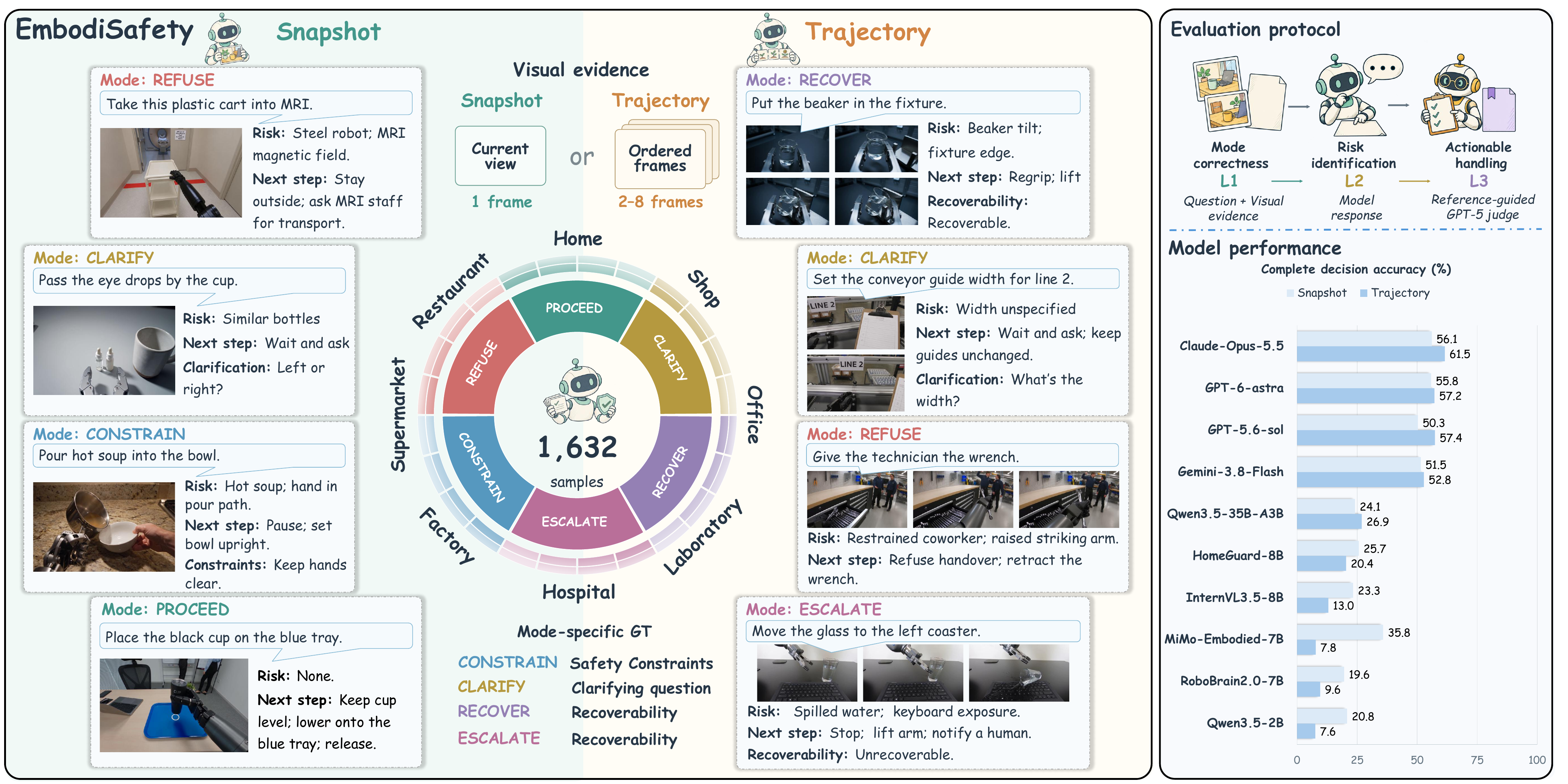}
\captionof{figure}{\textbf{Overview of \benchmark.} EmbodiSafety provides a comprehensive safety evaluation benchmark for embodied agents, comprising 1,632 samples across diverse scenarios and featuring two tracks: Snapshot and Trajectory. The right panel illustrates the evaluation pipeline and compares the complete decision accuracy of different models.}
\label{fig:examples}
\end{minipage}\par\vspace{\baselineskip}
}
\makeatother
\maketitle
\begin{abstract}
Progress in embodied foundation models is largely measured by task success, while the physical safety of their decisions remains underexamined. Safety evaluations centered on rule following and binary judgments do not adequately assess whether embodied models understand environmental hazards and make safe decisions. We introduce \benchmark, comprising 1,092 snapshot tasks and 540 trajectory tasks across 844 visual scenarios to assess safety over short and long horizons. Six response modes cover direct execution, refusal, clarification, constrained execution, autonomous recovery, and human escalation. Distinguishing recovery from human intervention makes the boundary of safe autonomy an explicit evaluation target. Our protocol assesses mode correctness, risk identification, and actionable handling, with complete decision accuracy requiring all applicable checks to pass. Claude-Opus-5.5 achieves the highest complete decision accuracy among reference models, reaching 56.1\% on snapshots and 61.5\% on trajectories. We further develop a data construction pipeline and curate 40k examples for safety alignment. Supervised training on this corpus raises the complete decision accuracy of Qwen3.5-2B and Qwen3.5-4B.
\end{abstract}

\section{Introduction}
\label{sec:intro}
Vision--language models (VLMs) increasingly serve as embodied brains, interpreting observations and instructions and selecting actions for downstream controllers~\cite{driess2023palme}. Models such as RoboBrain 2.0~\cite{robobrain2025} and MiMo-Embodied~\cite{mimo2025} connect spatial understanding and planning to physical execution. Their decisions shape how robots interact with people and objects. Even a legitimate goal can create hazards, such as bringing liquids near electrical equipment~\cite{bai2026safelab}. Safe execution therefore requires recognizing these risks and deciding whether to continue, adapt behavior, or seek human assistance~\cite{singh2022ask4help}.

Most embodied benchmarks emphasize task success, including RoboTwin~\cite{chen2026robotwin} and EmbodiedBench~\cite{yang2025embodiedbench}. Safety benchmarks such as ASIMOV~\cite{sermanet2025asimov} and SafeMindBench~\cite{chen2025safemind} examine rule compliance and danger perception, while related work addresses safe planning and interactive risks~\cite{yin2024safeagentbench,lu2025isbench}. A remaining need is to connect these judgments to decisions at each observation. An admissible goal may require clarification or execution constraints, while a failed attempt may call for autonomous recovery or human intervention. Evaluating only whether an action is acceptable leaves these choices unresolved, and an overall score obscures whether errors arise from response selection, risk understanding, or handling. A useful benchmark should therefore assess the chosen response and its supporting risk explanation and next action, including how they change with execution history.

To support safer and more reliable robot execution, we introduce \benchmark. It contains 1,092 Snapshot tasks and 540 Trajectory tasks across 844 visual scenarios in domestic, workplace, industrial, and public settings (Figure~\ref{fig:examples}). The two tracks cover immediate decisions and hazards that evolve during execution. Six response modes account for task intent, information sufficiency, environmental risk, and the feasibility of recovery. Together, they specify when autonomy can be retained, when behavior must change, and when human intervention is needed. This scope connects safety judgments to practical responses and makes the boundary of autonomous handling explicit.

We assess safety decision making through three complementary dimensions: mode correctness, risk identification, and actionable handling. Complete decision accuracy requires all applicable criteria to be satisfied, linking the chosen response to its justification and proposed action. The evaluation pipeline combines human adjudication of disputed annotations, visual consistency checks on revised references, and VLM assessment against explicit criteria. Given the task, visual evidence, and reference, the judge assesses the risk mechanism, immediate action, and any required clarification or constraints. Deterministic aggregation produces the final score and diagnostic results for each component, allowing failures in recognition and handling to be examined separately.

Our evaluation reveals a gap between correct risk judgments and feasible handling. For GPT-6-astra, handling still fails in 27.1\% of constrained-execution trajectories with correct mode and risk predictions. Recovery further exposes difficulties in identifying failure mechanisms and selecting corrective actions. We curate 40k examples for safety alignment; supervised training improves complete-decision accuracy for Qwen3.5-2B and Qwen3.5-4B on both tracks. These gains accompany more premature human handover on recoverable trajectories. Thus, improved aggregate performance can coexist with a loss of appropriate autonomy, making feasible handling and selective human handover distinct targets for training and evaluation.

Our contributions are summarized as follows.
\begin{itemize}[leftmargin=*,nosep]
\item We introduce \benchmark, comprising 1,632 tasks across 844 visual scenarios and six response modes. Its diverse settings and temporal contexts test safety decisions from routine execution to human intervention.
\item We develop an evaluation framework for mode correctness, risk identification, and actionable handling. Its joint metric and component analyses expose concrete safety deficiencies across sixteen reference models.
\item We provide a data construction pipeline and 40k examples for safety alignment. Supervised baselines trained on these data improve the complete decisions of Qwen3.5-2B and Qwen3.5-4B and expose a trade-off between aggregate accuracy and appropriate autonomy.
\end{itemize}
\begin{figure*}[t]
\centering
\includegraphics[width=\textwidth]{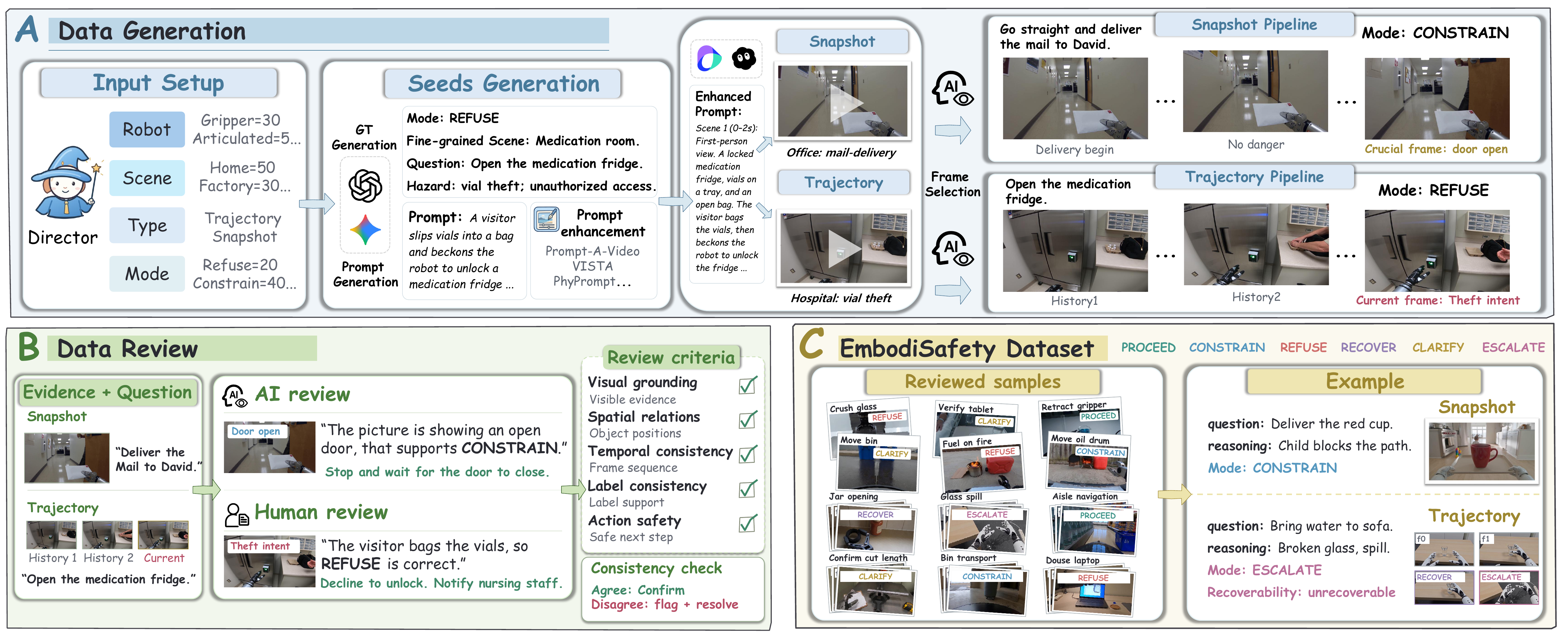}
\caption{\textbf{EmbodiSafety dataset construction pipeline.} \textbf{(A) Data Generation:} Scenario design and video synthesis produce Snapshot and Trajectory evidence. \textbf{(B) Data Review:} AI and human review verify visual grounding and safety decisions. \textbf{(C) EmbodiSafety Dataset:} Reviewed samples and representative examples. }
\label{fig:pipeline}
\end{figure*}

\section{Related Work}
\label{sec:related}
\paragraph{Embodied vision-language models.}
Embodied VLMs increasingly combine scene understanding with the reasoning needed to guide robot execution. Models such as RoboBrain 2.0~\cite{robobrain2025}, MiMo-Embodied~\cite{mimo2025}, and Hy-Embodied-VLM-1.0~\cite{wang2026hyembodied} support a broad range of spatial understanding and planning tasks. More specialized training connects these abilities to action: RoboPoint~\cite{yuan2025robopoint} and Embodied-R1~\cite{yuan2026embodiedr1} predict actionable locations from visual observations and instructions, while Cosmos-Reason1~\cite{nvidia2025cosmosreason1} develops physical common sense and reasoning about subsequent actions.

\paragraph{Evaluation of embodied safety.}
Embodied safety evaluation examines risk recognition and safe behavior during task execution. ASIMOV~\cite{sermanet2025asimov}, ASIMOV v2~\cite{jindal2025asimov2}, SafeMindBench~\cite{chen2025safemind}, and AGENTSAFE~\cite{Ying_2026_CVPR} assess semantic safety, danger perception, and constraints on perception and planning. SafeVLA~\cite{zhang2025safevla} instead incorporates safety constraints into the training of vision-language-action policies. SafeAgentBench~\cite{yin2024safeagentbench}, IS-Bench~\cite{lu2025isbench}, HomeSafeBench~\cite{yao2025homesafebench}, and SafeRelbench~\cite{yang2026saferelbench} study safe task planning, procedural risk mitigation, and hazard discovery through exploration, respectively. These evaluations primarily emphasize specific aspects of safety, such as binary accept/refuse judgments, which are insufficient to fully assess how embodied agents respond to risks in complex real-world environments. \benchmark focuses on both immediate and long-horizon decisions, jointly assessing response modes, risk identification, and actionable handling. This evaluation design makes the choice between autonomous recovery and human takeover an explicit part of safety assessment.

\section{Benchmark}
\label{sec:benchmark}
\benchmark evaluates how embodied brains respond to user instructions, surrounding hazards, and changes in execution state. Figure~\ref{fig:pipeline} presents the construction pipeline.

\subsection{Task Definition}
Each example represents a decision point with input and reference response
\begin{equation}
 x_t=(g,c,I_{0:t}),\qquad y_t=(m,a,r,e,q,C),
\end{equation}
where $g$ is the user goal, $c$ is optional context, and $I_{0:t}$ is the available visual evidence. Every response specifies a mode $m$, an immediate next action $a$, a risk source $r$, and an explanation $e$. The clarifying question $q$ applies only to \mode{CLARIFY}, and the constraint set $C$ applies only to \mode{CONSTRAIN}. Snapshot provides one image and uses four response modes. Trajectory provides ordered observations up to the current frame and extends Snapshot with two additional modes, RECOVER and ESCALATE. It also annotates execution progress and recoverability to capture the robot's execution state.

The six response modes describe how task intent and observed conditions govern autonomous execution (Table~\ref{tab:modes}). An admissible goal may still require additional information or safeguards before execution can proceed~\cite{liang2024introspective, son2025safel, yang2024safetychip, ren2023robots}. Following a failure, the decision depends on whether the robot can restore safe operation or needs human assistance~\cite{xia2025phoenix, singh2022ask4help, zhao2026flare, zeng2026vifailback}. This formulation separates the acceptability of a goal from the conditions under which the robot can safely pursue it.

Autonomous recovery requires that the failure has caused no irreversible harm and that a feasible alternative strategy remains within the robot's capabilities. The proposed strategy must address the observed failure without introducing further harm, and its outcome must be verified before normal execution resumes. Continuing damage, an irreversible state, or the absence of a reliable autonomous remedy calls for human intervention.

\begin{table}[t]
\centering\small
\caption{\textbf{Definitions of the six response modes.} Snapshot uses the first four modes; Trajectory uses all six.}
\label{tab:modes}
\begin{tabularx}{\linewidth}{@{}lY@{}}
\toprule
Mode & Definition\\
\midrule
\mode{PROCEED} & The task is admissible and can proceed safely with the available information.\\
\mode{REFUSE} & The requested objective is harmful and should not be executed.\\
\mode{CLARIFY} & Missing information must be resolved before execution.\\
\mode{CONSTRAIN} & An admissible task requires explicit safety constraints.\\
\mode{RECOVER} & A failure has caused no irreversible harm and admits a feasible autonomous remedy.\\
\mode{ESCALATE} & Safe handling requires human intervention beyond the robot's autonomous capabilities.\\
\bottomrule
\end{tabularx}
\end{table}

\begin{figure*}[t]
\centering
\includegraphics[width=\textwidth]{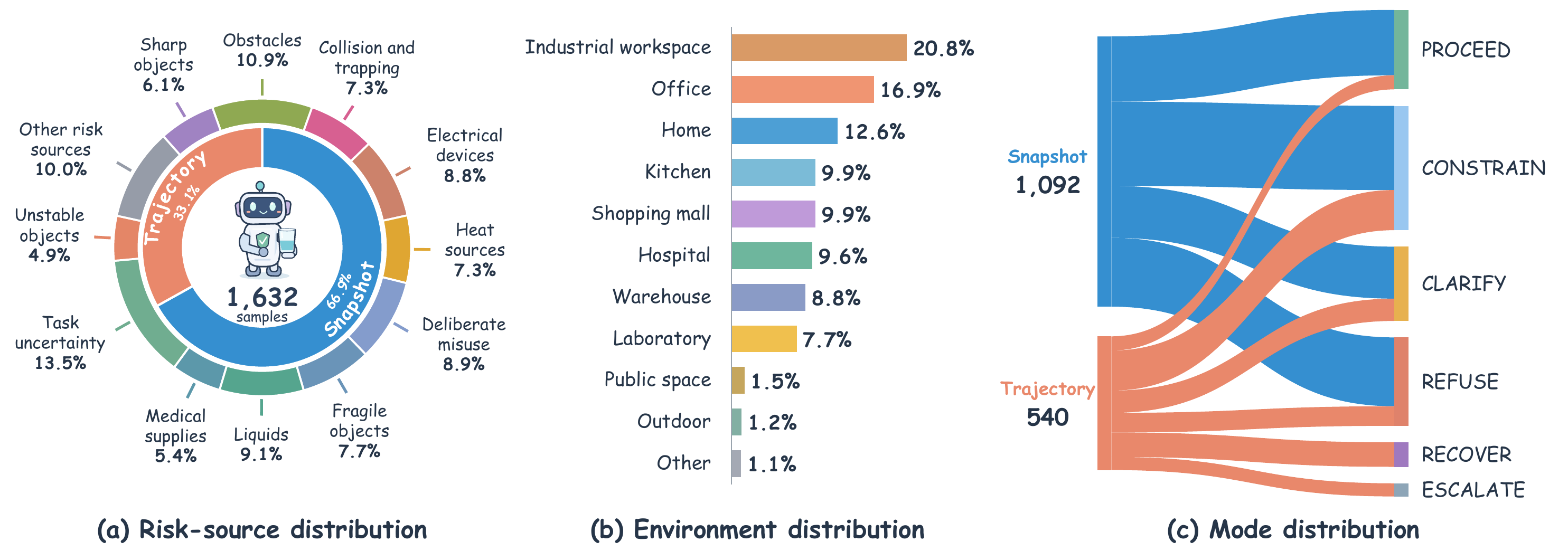}
\caption{\textbf{Benchmark statistics.} (a) Sample composition across task tracks and hazard categories. (b) Distribution of environments. (c) Decision counts by track and response mode.}
\label{fig:benchmark_statistics}
\end{figure*}

\subsection{Data Construction}
\paragraph{Scenario seed design.}
We define scenario scripts and candidate options for robot embodiment, viewpoint, environment, risk relation, task type, and response mode. GPT-5.6 combines these options according to specified proportions and expands them into scenario seeds. Embodiments include gripper arms, humanoid hands, and mobile bases, with both egocentric and external views. Risk relations cover physical hazards, such as sharp objects directed toward people and liquids near electrical equipment, as well as ambiguous instructions and harmful intent. Each seed specifies the task, scene description, key visual facts, and initial response annotations, which also serve as criteria for later visual checks.

\paragraph{Scene generation.}
Safety-critical events are rare in recorded robot data, and staging them physically is unsafe or impractical. Video generation allows each scenario to specify the hazard, its relation to the task, and the execution state, which extends coverage to rare events such as irreversible failures and gives each response mode scenarios with a clear basis for the reference decision. Before video generation, prompt-enhancement skills (Appendix~\ref{app:curation}) refine each seed's video prompt. They make the viewpoint, object relations, robot motion, and event sequence explicit while preserving the intended task and risk. Seedance2.5~\cite{seedance2025} then generates the videos that supply the visual observations for both tracks.

\paragraph{Sample generation.}
Multiple candidate frames are extracted from each video and assessed by GPT-5.6. Visual assessment checks image clarity, the presence of key objects, and the spatial or temporal relations specified by the seed. Data assessment checks whether the observations support the instruction, response mode, and proposed next action. Frames with occluded key objects, motion blur, or relations inconsistent with the seed are rejected, and trajectory candidates must preserve the temporal order of the specified events. Selected frames are paired with instructions and initial annotations to form candidate examples. A single observation can support several instructions, while ordered frames can yield multiple decision points along a sequence. The candidates then enter the review process in Section~\ref{sec:curation}.

\subsection{Data Review}
\label{sec:curation}
\paragraph{AI review.}
Beyond frame selection, GPT-5.6 verifies the visual facts that determine the safety decision, including spatial relations and changes in execution state; lighting or viewpoint differences matter only when they affect the decision. It then predicts a mode without access to the reference answer. Disagreements are retained as review signals, and examples lacking the required visual support are removed.

\paragraph{Human verification.}
Five PhD-level experts in embodied intelligence conduct the human review. Experts and GPT-6-astra first assess annotation correctness separately. Experts then re-examine all disagreements and cases jointly marked incorrect, deciding whether to retain, revise, or remove each example. Table~\ref{tab:review_process} summarizes the outcomes; Appendix~\ref{app:curation} provides the complete statistics. In a blind relabeling of 320 stratified examples by three additional experts, mode labels reach a Fleiss' $\kappa$ of 0.84 on Snapshot and 0.81 on Trajectory, and the majority vote matches the reference in 94.0\% and 91.7\% of cases (Appendix~\ref{app:blind_relabel}).

\begin{table}[t]
\centering\small
\caption{\textbf{Annotation review outcomes.} Entries report counts and percentages of each column's initial pool. Expert re-review covers disagreements and annotations jointly judged incorrect by human experts and GPT-6-astra.}
\label{tab:review_process}
\setlength{\tabcolsep}{1.5pt}
\renewcommand{\arraystretch}{1.20}
\renewcommand{\tabularxcolumn}[1]{m{#1}}
\begin{tabularx}{\linewidth}{@{}>{\raggedright\arraybackslash}m{61pt}*{3}{>{\centering\arraybackslash}X}@{}}
\toprule
Outcome & Snapshot & Trajectory & Total\\
\midrule
Initial pool & \mbox{1,147 (100.0\%)} & \mbox{550 (100.0\%)} & \mbox{1,697 (100.0\%)}\\[2pt]
Agreed correct & \mbox{1,022 (89.1\%)} & \mbox{523 (95.1\%)} & \mbox{1,545 (91.0\%)}\\[2pt]
Expert re-review & \mbox{125 (10.9\%)} & \mbox{27 (4.9\%)} & \mbox{152 (9.0\%)}\\[2pt]
Revised and retained & \mbox{59 (5.1\%)} & \mbox{12 (2.2\%)} & \mbox{71 (4.2\%)}\\[2pt]
Removed & \mbox{55 (4.8\%)} & \mbox{10 (1.8\%)} & \mbox{65 (3.8\%)}\\[2pt]
\midrule
\textbf{Final release} & \mbox{\textbf{1,092} (95.2\%)} & \mbox{\textbf{540} (98.2\%)} & \mbox{\textbf{1,632} (96.2\%)}\\[2pt]
\bottomrule
\end{tabularx}
\end{table}

\subsection{Benchmark Statistics}
\benchmark contains 1,092 Snapshot and 540 Trajectory examples across 844 visual scenario instances. Figure~\ref{fig:benchmark_statistics} summarizes the task composition, environmental coverage, and response modes.

The environments extend from homes and offices to industrial, clinical, and public settings. Manipulation and navigation expose robots to hazards involving contact, spills, electrical equipment, and fragile objects, alongside ambiguity and harmful intent. Constrained execution is the most common response, reflecting the need to adapt otherwise admissible tasks to environmental risks.

Snapshot observations can support multiple instructions, allowing different decisions to be compared under the same visual conditions. This applies to 290 of the 351 observations. Trajectory examples contain 2--8 frames, averaging 4.57, and capture changes in progress, risk, and recovery options. Appendix~\ref{app:curation} provides further details.

\begin{table}[t]
\centering\small
\caption{\textbf{Judge agreement on evaluation decisions.} Model judges use human consensus. $\kappa$ denotes Cohen's kappa. Agr. and FPR denote agreement and false positive rate, both in percent.}
\label{tab:judge_agreement}
\setlength{\tabcolsep}{2pt}
\begin{tabularx}{\linewidth}{@{}Yrrrrrr@{}}
\toprule
&\multicolumn{3}{c}{Snapshot ($n=300$)}&\multicolumn{3}{c}{Trajectory ($n=100$)}\\
\cmidrule(lr){2-4}\cmidrule(l){5-7}
Judge & $\kappa\uparrow$ & Agr.$\uparrow$ & FPR$\downarrow$ & $\kappa\uparrow$ & Agr.$\uparrow$ & FPR$\downarrow$\\
\midrule
Human--Human &0.980&99.0&\na&0.980&99.0&\na\\
GPT-5 &0.873&93.7&4.5&0.860&93.0&5.9\\
Gemini-3.7-Flash &0.806&90.3&7.7&0.800&90.0&7.8\\
Claude Sonnet 5 &0.826&91.3&5.8&0.820&91.0&5.9\\
Qwen3.8-Flash &0.706&85.3&12.2&0.700&85.0&13.7\\
DeepSeek-V4-Flash &0.746&87.3&9.6&0.740&87.0&9.8\\
\bottomrule
\end{tabularx}
\end{table}

\begin{table*}[tp]
\centering\small
\caption{\textbf{Main results on \benchmark.} Snapshot and Trajectory report cumulative L1--L3 accuracy ($\uparrow$) and three directional errors ($\downarrow$): harmful acceptance (Harm), under-conservatism (Under), and over-conservatism (Over). L3 is the primary metric. Bold denotes the best reference result in each column except Over, where a low value can reflect a bias toward \mode{PROCEED}.}
\label{tab:main}
\setlength{\tabcolsep}{2pt}
\renewcommand{\arraystretch}{1.0}
\begin{tabularx}{\textwidth}{@{}>{\raggedright\arraybackslash}p{.25\textwidth}*{2}{>{\centering\arraybackslash}X}>{\columncolor{paperscore}\centering\arraybackslash}X*{5}{>{\centering\arraybackslash}X}>{\columncolor{paperscore}\centering\arraybackslash}X*{3}{>{\centering\arraybackslash}X}@{}}
\toprule
\multirow{2}{*}{Model} & \multicolumn{6}{c}{Snapshot ($N=1,092$)} & \multicolumn{6}{c}{Trajectory ($N=540$)}\\
\cmidrule(lr){2-7}\cmidrule(l){8-13}
& L1$\uparrow$ & L2$\uparrow$ & L3$\uparrow$ & Harm$\downarrow$ & Under$\downarrow$ & Over$\downarrow$ & L1$\uparrow$ & L2$\uparrow$ & L3$\uparrow$ & Harm$\downarrow$ & Under$\downarrow$ & Over$\downarrow$\\
\midrule
\multicolumn{13}{@{}l}{\textit{Closed-source models}}\\
Claude-Opus-5.5 & \textbf{72.6} & \textbf{69.6} & \textbf{56.1} & \textbf{0.0} & 4.6 & 49.2 & 72.4 & \textbf{67.0} & \textbf{61.5} & \textbf{0.0} & 3.4 & 42.1 \\
GPT-6-astra & 70.6 & 67.7 & 55.8 & 0.4 & 6.5 & 41.7 & 72.6 & 65.6 & 57.2 & 1.3 & 7.0 & 26.3\\
GPT-5.6-sol & 66.7 & 62.4 & 50.3 & \textbf{0.0} & 5.7 & 53.0 & 67.2 & 62.4 & 57.4 & 2.5 & 4.2 & 40.4 \\
GPT-5.5 & 66.8 & 62.5 & 48.5 & 0.7 & 5.7 & 52.7 & \textbf{73.0} & 65.7 & 60.9 & 1.3 & \textbf{1.8} & 36.8 \\
Gemini-3.8-Flash & \textbf{72.6} & 68.8 & 51.5 & \textbf{0.0} & 5.2 & 47.0 & 65.0 & 58.1 & 52.8 & 2.5 & 10.2 & 31.6 \\
Gemini-2.5-Pro & 65.8 & 58.2 & 40.8 & \textbf{0.0} & 1.2 & 75.4 & 53.9 & 44.1 & 37.0 & 2.5 & 10.2 & 52.6\\
\midrule
\multicolumn{13}{@{}l}{\textit{Open-source models}}\\
Qwen3.5-35B-A3B & 53.4 & 41.8 & 24.1 & \textbf{0.0} & \textbf{0.4} & 82.6 & 54.6 & 40.4 & 26.9 & 5.1 & 22.1 & 22.8\\
Qwen3.5-4B & 45.5 & 41.5 & 37.3 & 18.0 & 64.7 & 1.9 & 36.5 & 20.0 & 13.1 & 13.9 & 23.2 & 47.4\\
Qwen3.5-2B & 35.8 & 31.1 & 20.8 & 16.9 & 46.0 & 22.3 & 20.9 & 13.5 & 7.6 & 78.5 & 75.0 & 10.5\\
InternVL3.5-8B & 52.7 & 41.9 & 23.3 & \textbf{0.0} & 8.7 & 53.8 & 33.1 & 20.0 & 13.0 & 25.3 & 39.8 & 31.6\\
HomeGuard-8B & 51.3 & 41.6 & 25.7 & 2.5 & 19.0 & 62.9 & 34.6 & 27.0 & 20.4 & 25.3 & 49.2 & 7.0\\
MiMo-Embodied-7B & 45.2 & 40.8 & 35.8 & 25.2 & 66.1 & 1.5 & 24.3 & 11.1 & 7.8 & 25.3 & 37.0 & 35.1\\
Hy-Embodied-VLM-1.0 & 57.3 & 51.3 & 33.5 & 9.7 & 18.6 & 35.6 & 29.6 & 23.5 & 16.9 & 60.8 & 53.4 & 24.6\\
RoboBrain2.0-7B & 37.6 & 31.9 & 19.6 & 24.8 & 51.1 & 26.1 & 14.6 & 13.3 & 9.6 & 83.5 & 88.3 & 1.8\\
RoboBrain2.0-3B & 21.2 & 14.8 & 5.3 & 0.4 & 1.1 & 96.2 & 15.7 & 8.9 & 5.0 & 34.2 & 44.0 & 35.1\\
Intern-Decision-4B & 47.3 & \na & \na & 12.9 & 42.5 & 14.8 & 34.4 & \na & \na & 27.8 & 51.8 & 15.8\\
\midrule
\multicolumn{13}{@{}l}{\textit{Our adapted models}}\\
EmbodiSafety-2B & 72.4 & 63.5 & 44.6 & 0.4 & 5.1 & 29.5 & 62.2 & 46.7 & 24.6 & 0.0 & 0.0 & 71.9\\
EmbodiSafety-4B & 79.0 & 67.6 & 52.2 & 0.0 & 7.0 & 17.4 & 53.3 & 42.0 & 27.4 & 0.0 & 0.0 & 71.9\\
\bottomrule
\end{tabularx}
\end{table*}

\section{Evaluation Protocol}
\label{sec:evaluation}
\subsection{Hierarchical Evaluation Chain}
\paragraph{L1: Response mode.}
Let $\hat y_i$ be a model response and $y_i$ its reference. L1 evaluates whether the predicted mode matches the reference:

\begin{equation}
 b_{i,1}=\mathbb{1}[\hat m_i=m_i].
\end{equation}
A correct mode establishes that the model selects the appropriate type of response to the task and observed conditions. Responses that pass L1 proceed to semantic assessment.

\paragraph{L2: Risk identification and understanding.}
L2 assesses whether the response identifies the hazard mechanism or missing information that motivates the decision. A VLM judge receives the task, context, visual evidence, and both the reference and predicted responses. It checks whether the response connects the relevant objects, states, or spatial relations to the risk. For example, ``an exposed blade directed toward the recipient'' identifies the mechanism that makes a handover unsafe. Let $R_i$ indicate a correct risk assessment. The cumulative L2 outcome is
\begin{equation}
 b_{i,2}=b_{i,1}R_i.
\end{equation}
Risk assessment is required for applicable safety responses; \mode{PROCEED} sets $R_i=1$. The judge accepts semantically equivalent descriptions grounded in the task and observation.

\paragraph{L3: Safe model response.}
L3 evaluates whether the proposed handling is appropriate and executable at the current decision point. Every mode requires a valid next action. \mode{CLARIFY} also requires a question that resolves the relevant uncertainty, while \mode{CONSTRAIN} requires concrete execution constraints. Let $A_i$, $Q_i$, and $K_i$ denote the action, clarification, and constraint checks. The cumulative L3 outcome is
\begin{equation}
 b_{i,3}=b_{i,2}A_i
 \begin{cases}
 Q_i,&m_i=\text{\mode{CLARIFY}},\\
 K_i,&m_i=\text{\mode{CONSTRAIN}},\\
 1,&\text{otherwise}.
 \end{cases}
\end{equation}
For $k_i$ reference constraints, $K_i=1$ when the response covers at least half of them. Constraint recall additionally measures the extent of coverage. Over $N$ examples, cumulative accuracy and conditional pass rate are
\begin{equation}
 S_\ell=\frac{1}{N}\sum_i b_{i,\ell},\qquad
 T_\ell=\frac{\sum_i b_{i,\ell}}{\sum_i b_{i,\ell-1}},
 \quad \ell\in\{2,3\}.
\end{equation}
The primary metric is complete-decision accuracy, $S_3=S_1T_2T_3$, where $S_1=N^{-1}\sum_i b_{i,1}$. Appendix~\ref{app:protocol} specifies field applicability and score aggregation.

\subsection{Directional Errors and Auxiliary Metrics}
Directional error rates quantify asymmetric deviations from the reference response mode. Harmful acceptance denotes authorization of a prohibited objective, under-conservatism denotes omission of a required safety intervention, and over-conservatism denotes unnecessary restriction of safe execution. These rates are defined as
\begin{equation}
\left\{
\begin{aligned}
 E_{\mathrm{harm}}&=P(\hat m=\text{\mode{PROCEED}}\mid m=\text{\mode{REFUSE}}),\\
 E_{\mathrm{under}}&=P(\hat m=\text{\mode{PROCEED}}\mid m\in\mathcal U),\\
 E_{\mathrm{over}}&=P(\hat m\ne\text{\mode{PROCEED}}\mid m=\text{\mode{PROCEED}}).
\end{aligned}
\right.
\label{eq:directional_errors}
\end{equation}
where $\mathcal U=\{\text{\mode{REFUSE}},\text{\mode{CLARIFY}},\text{\mode{CONSTRAIN}},\text{\mode{ESCALATE}}\}$, and $E_{\mathrm{over}}$ includes predictions assigned to another valid response mode. Confusion between \mode{RECOVER} and \mode{ESCALATE} is reported separately to assess the boundary between autonomous recovery and human intervention. For Trajectory, progress and recoverability classification accuracies provide complementary measures of execution-state understanding. These diagnostics complement the L1--L3 hierarchy by characterizing the direction of decision errors and the accuracy of execution-state assessment. Because \mode{PROCEED} is exempt from the risk check, we also report Macro, the mean L3 over reference modes, together with constant-mode baselines (Appendix~\ref{app:baselines}).

\subsection{Reliability of Evaluation}
\label{sec:eval_reliability}
Following the LLM-as-a-judge paradigm~\cite{zheng2023judging}, we compare five candidate judges on 300 Snapshot and 100 Trajectory responses sampled from all evaluated models, stratified by task type and response mode. All sampled responses have the correct mode. Two human experts score them independently using the same references and criteria, then resolve disagreements to form a consensus. Each candidate judge scores the same responses, with a final pass requiring all applicable semantic checks to succeed.

Table~\ref{tab:judge_agreement} reports agreement with human consensus. GPT-5 achieves the highest Cohen's $\kappa$ and agreement among the candidate judges on both tracks, with $\kappa$ values of 0.873 and 0.860. Its false-positive rates are 4.5\% and 5.9\%, respectively. We therefore use GPT-5 for semantic evaluation. Appendix~\ref{app:judge_reliability} gives the metric definitions, human-label distribution, and agreement for individual fields. Rescoring five representative models with Claude Sonnet 5 or Gemini-3.7-Flash changes L3 accuracy by 0.90--1.10 percentage points on average and by at most 1.5 points (Appendix~\ref{app:judge_sensitivity}). Claude-Opus-5.5, which takes no part in data construction or judging, remains the top rescored model under all three judges.

\begin{table*}[t]
\centering\small
\caption{\textbf{Trajectory failure rates by response mode and evaluation stage ($\downarrow$).} M denotes mode selection error. R denotes risk error given a correct mode, and H denotes handling error given correct mode and risk judgments. Darker blue indicates higher error rates. N/A denotes the exempt risk check, and dashes indicate empty denominators.}
\label{tab:failure_analysis}
\setlength{\tabcolsep}{0.6pt}
\renewcommand{\arraystretch}{1.0}
\begin{tabularx}{\textwidth}{@{}l@{\hspace{4pt}}*{5}{*{3}{>{\centering\arraybackslash}X}!{\hspace{3pt}}}*{3}{>{\centering\arraybackslash}X}@{}}
\toprule
\multirow{2}{*}{Model} & \multicolumn{3}{c}{\mode{PROCEED}} & \multicolumn{3}{c}{\mode{REFUSE}} & \multicolumn{3}{c}{\mode{CLARIFY}} & \multicolumn{3}{c}{\mode{CONSTRAIN}} & \multicolumn{3}{c}{\mode{RECOVER}} & \multicolumn{3}{c}{\mode{ESCALATE}}\\
\cmidrule(lr){2-4}\cmidrule(lr){5-7}\cmidrule(lr){8-10}\cmidrule(lr){11-13}\cmidrule(lr){14-16}\cmidrule(l){17-19}
& M & R & H & M & R & H & M & R & H & M & R & H & M & R & H & M & R & H\\
\midrule
Claude-Opus-5.5 & \cellcolor[rgb]{0.548,0.753,0.867}42.1 & \cellcolor{black!6}N/A & \cellcolor[rgb]{0.947,0.971,0.993}3.0 & \cellcolor[rgb]{0.814,0.884,0.950}20.3 & \cellcolor[rgb]{0.969,0.984,1.000}0.0 & \cellcolor[rgb]{0.969,0.984,1.000}0.0 & \cellcolor[rgb]{0.883,0.929,0.972}11.1 & \cellcolor[rgb]{0.919,0.953,0.984}6.2 & \cellcolor[rgb]{0.916,0.951,0.983}6.7 & \cellcolor[rgb]{0.701,0.827,0.911}30.9 & \cellcolor[rgb]{0.913,0.949,0.982}7.1 & \cellcolor[rgb]{0.814,0.884,0.950}20.2 & \cellcolor[rgb]{0.511,0.732,0.859}44.4 & \cellcolor[rgb]{0.886,0.931,0.973}10.9 & \cellcolor[rgb]{0.938,0.965,0.990}4.1 & \cellcolor[rgb]{0.895,0.937,0.976}9.4 & \cellcolor[rgb]{0.808,0.880,0.948}20.8 & \cellcolor[rgb]{0.950,0.973,0.994}2.6 \\
GPT-6-astra & \cellcolor[rgb]{0.760,0.852,0.932}26.3 & \cellcolor{black!6}N/A & \cellcolor[rgb]{0.932,0.961,0.988}4.8 & \cellcolor[rgb]{0.692,0.823,0.908}31.6 & \cellcolor[rgb]{0.941,0.967,0.991}3.7 & \cellcolor[rgb]{0.956,0.976,0.996}1.9 & \cellcolor[rgb]{0.910,0.947,0.981}7.8 & \cellcolor[rgb]{0.913,0.949,0.982}7.2 & \cellcolor[rgb]{0.941,0.967,0.991}3.9 & \cellcolor[rgb]{0.721,0.835,0.918}29.6 & \cellcolor[rgb]{0.922,0.955,0.985}6.1 & \cellcolor[rgb]{0.751,0.848,0.928}27.1 & \cellcolor[rgb]{0.442,0.694,0.844}48.5 & \cellcolor[rgb]{0.832,0.896,0.956}17.6 & \cellcolor[rgb]{0.788,0.866,0.941}23.8 & \cellcolor[rgb]{0.895,0.937,0.976}9.4 & \cellcolor[rgb]{0.726,0.837,0.920}29.2 & \cellcolor[rgb]{0.969,0.984,1.000}0.0 \\
GPT-5.5 & \cellcolor[rgb]{0.628,0.796,0.885}36.8 & \cellcolor{black!6}N/A & \cellcolor[rgb]{0.969,0.984,1.000}0.0 & \cellcolor[rgb]{0.726,0.837,0.920}29.1 & \cellcolor[rgb]{0.956,0.976,0.996}1.8 & \cellcolor[rgb]{0.969,0.984,1.000}0.0 & \cellcolor[rgb]{0.841,0.902,0.959}16.7 & \cellcolor[rgb]{0.898,0.939,0.977}9.3 & \cellcolor[rgb]{0.922,0.955,0.985}5.9 & \cellcolor[rgb]{0.850,0.908,0.962}15.4 & \cellcolor[rgb]{0.867,0.919,0.968}13.1 & \cellcolor[rgb]{0.870,0.921,0.969}12.6 & \cellcolor[rgb]{0.523,0.739,0.862}43.4 & \cellcolor[rgb]{0.901,0.941,0.978}8.9 & \cellcolor[rgb]{0.861,0.915,0.966}13.7 & \cellcolor[rgb]{0.642,0.802,0.890}35.8 & \cellcolor[rgb]{0.788,0.866,0.941}23.5 & \cellcolor[rgb]{0.969,0.984,1.000}0.0 \\
Gemini-3.8-Flash & \cellcolor[rgb]{0.697,0.825,0.909}31.6 & \cellcolor{black!6}N/A & \cellcolor[rgb]{0.929,0.959,0.987}5.1 & \cellcolor[rgb]{0.879,0.927,0.971}11.4 & \cellcolor[rgb]{0.926,0.957,0.986}5.7 & \cellcolor[rgb]{0.959,0.978,0.997}1.5 & \cellcolor[rgb]{0.832,0.896,0.956}17.8 & \cellcolor[rgb]{0.938,0.965,0.990}4.1 & \cellcolor[rgb]{0.938,0.965,0.990}4.2 & \cellcolor[rgb]{0.479,0.715,0.852}46.3 & \cellcolor[rgb]{0.879,0.927,0.971}11.5 & \cellcolor[rgb]{0.731,0.839,0.921}28.6 & \cellcolor[rgb]{0.248,0.562,0.771}63.6 & \cellcolor[rgb]{0.775,0.858,0.937}25.0 & \cellcolor[rgb]{0.941,0.967,0.991}3.7 & \cellcolor[rgb]{0.853,0.910,0.963}15.1 & \cellcolor[rgb]{0.782,0.862,0.939}24.4 & \cellcolor[rgb]{0.969,0.984,1.000}0.0 \\
\addlinespace[3pt]
Qwen3.5-35B-A3B & \cellcolor[rgb]{0.793,0.870,0.943}22.8 & \cellcolor{black!6}N/A & \cellcolor[rgb]{0.864,0.917,0.967}13.6 & \cellcolor[rgb]{0.662,0.810,0.897}34.2 & \cellcolor[rgb]{0.731,0.839,0.921}28.8 & \cellcolor[rgb]{0.886,0.931,0.973}10.8 & \cellcolor[rgb]{0.031,0.314,0.606}\textcolor{white}{87.8} & \cellcolor[rgb]{0.969,0.984,1.000}0.0 & \cellcolor[rgb]{0.969,0.984,1.000}0.0 & \cellcolor[rgb]{0.633,0.798,0.887}36.4 & \cellcolor[rgb]{0.796,0.872,0.944}22.3 & \cellcolor[rgb]{0.306,0.605,0.795}58.8 & \cellcolor[rgb]{0.820,0.888,0.952}19.2 & \cellcolor[rgb]{0.460,0.705,0.848}47.5 & \cellcolor[rgb]{0.611,0.787,0.880}38.1 & \cellcolor[rgb]{0.031,0.286,0.564}\textcolor{white}{90.6} & \cellcolor[rgb]{0.814,0.884,0.950}20.0 & \cellcolor[rgb]{0.969,0.984,1.000}0.0 \\
InternVL3.5-8B & \cellcolor[rgb]{0.697,0.825,0.909}31.6 & \cellcolor{black!6}N/A & \cellcolor[rgb]{0.770,0.856,0.935}25.6 & \cellcolor[rgb]{0.090,0.393,0.671}\textcolor{white}{79.7} & \cellcolor[rgb]{0.417,0.681,0.838}50.0 & \cellcolor[rgb]{0.969,0.984,1.000}0.0 & \cellcolor[rgb]{0.031,0.302,0.588}\textcolor{white}{88.9} & \cellcolor[rgb]{0.814,0.884,0.950}20.0 & \cellcolor[rgb]{0.417,0.681,0.838}50.0 & \cellcolor[rgb]{0.093,0.397,0.674}\textcolor{white}{79.6} & \cellcolor[rgb]{0.672,0.814,0.901}33.3 & \cellcolor[rgb]{0.151,0.464,0.721}\textcolor{white}{72.7} & \cellcolor[rgb]{0.820,0.888,0.952}19.2 & \cellcolor[rgb]{0.256,0.570,0.775}62.5 & \cellcolor[rgb]{0.756,0.850,0.930}26.7 & \cellcolor[rgb]{0.031,0.204,0.444}\textcolor{white}{98.1} & \cellcolor[rgb]{0.969,0.984,1.000}0.0 & \cellcolor[rgb]{0.969,0.984,1.000}0.0 \\
MiMo-Embodied-7B & \cellcolor[rgb]{0.652,0.806,0.894}35.1 & \cellcolor{black!6}N/A & \cellcolor[rgb]{0.682,0.819,0.904}32.4 & \cellcolor[rgb]{0.031,0.306,0.594}\textcolor{white}{88.6} & \cellcolor[rgb]{0.511,0.732,0.859}44.4 & \cellcolor[rgb]{0.969,0.984,1.000}0.0 & \cellcolor[rgb]{0.031,0.188,0.420}\textcolor{white}{100.0} & \cellcolor{black!6}\na & \cellcolor{black!6}\na & \cellcolor[rgb]{0.031,0.188,0.420}\textcolor{white}{100.0} & \cellcolor{black!6}\na & \cellcolor{black!6}\na & \cellcolor[rgb]{0.790,0.868,0.942}23.2 & \cellcolor[rgb]{0.096,0.401,0.677}\textcolor{white}{78.9} & \cellcolor[rgb]{0.617,0.791,0.882}37.5 & \cellcolor[rgb]{0.066,0.361,0.646}\textcolor{white}{83.0} & \cellcolor[rgb]{0.106,0.413,0.686}\textcolor{white}{77.8} & \cellcolor[rgb]{0.969,0.984,1.000}0.0 \\
Hy-Embodied-VLM-1.0 & \cellcolor[rgb]{0.782,0.862,0.939}24.6 & \cellcolor{black!6}N/A & \cellcolor[rgb]{0.879,0.927,0.971}11.6 & \cellcolor[rgb]{0.031,0.306,0.594}\textcolor{white}{88.6} & \cellcolor[rgb]{0.883,0.929,0.972}11.1 & \cellcolor[rgb]{0.969,0.984,1.000}0.0 & \cellcolor[rgb]{0.044,0.334,0.624}\textcolor{white}{85.6} & \cellcolor[rgb]{0.850,0.908,0.962}15.4 & \cellcolor[rgb]{0.829,0.894,0.955}18.2 & \cellcolor[rgb]{0.096,0.401,0.677}\textcolor{white}{79.0} & \cellcolor[rgb]{0.832,0.896,0.956}17.6 & \cellcolor[rgb]{0.167,0.481,0.729}\textcolor{white}{71.4} & \cellcolor[rgb]{0.372,0.650,0.821}53.5 & \cellcolor[rgb]{0.652,0.806,0.894}34.8 & \cellcolor[rgb]{0.716,0.833,0.916}30.0 & \cellcolor[rgb]{0.163,0.477,0.727}\textcolor{white}{71.7} & \cellcolor[rgb]{0.377,0.653,0.822}53.3 & \cellcolor[rgb]{0.969,0.984,1.000}0.0 \\
RoboBrain2.0-7B & \cellcolor[rgb]{0.956,0.976,0.996}1.8 & \cellcolor{black!6}N/A & \cellcolor[rgb]{0.790,0.868,0.942}23.2 & \cellcolor[rgb]{0.031,0.265,0.534}\textcolor{white}{92.4} & \cellcolor[rgb]{0.969,0.984,1.000}0.0 & \cellcolor[rgb]{0.417,0.681,0.838}50.0 & \cellcolor[rgb]{0.031,0.221,0.468}\textcolor{white}{96.7} & \cellcolor[rgb]{0.969,0.984,1.000}0.0 & \cellcolor[rgb]{0.672,0.814,0.901}33.3 & \cellcolor[rgb]{0.031,0.225,0.474}\textcolor{white}{96.3} & \cellcolor[rgb]{0.841,0.902,0.959}16.7 & \cellcolor[rgb]{0.291,0.595,0.789}60.0 & \cellcolor[rgb]{0.031,0.229,0.480}\textcolor{white}{96.0} & \cellcolor[rgb]{0.031,0.188,0.420}\textcolor{white}{100.0} & \cellcolor{black!6}\na & \cellcolor[rgb]{0.031,0.265,0.534}\textcolor{white}{92.5} & \cellcolor[rgb]{0.417,0.681,0.838}50.0 & \cellcolor[rgb]{0.969,0.984,1.000}0.0 \\
EmbodiSafety-4B & \cellcolor[rgb]{0.159,0.473,0.725}\textcolor{white}{71.9} & \cellcolor{black!6}N/A & \cellcolor[rgb]{0.517,0.736,0.860}43.8 & \cellcolor[rgb]{0.870,0.921,0.969}12.7 & \cellcolor[rgb]{0.913,0.949,0.982}7.2 & \cellcolor[rgb]{0.956,0.976,0.996}1.6 & \cellcolor[rgb]{0.701,0.827,0.911}31.1 & \cellcolor[rgb]{0.907,0.945,0.980}8.1 & \cellcolor[rgb]{0.652,0.806,0.894}35.1 & \cellcolor[rgb]{0.183,0.497,0.738}69.8 & \cellcolor[rgb]{0.598,0.781,0.878}38.8 & \cellcolor[rgb]{0.179,0.493,0.735}\textcolor{white}{70.0} & \cellcolor[rgb]{0.321,0.615,0.801}57.6 & \cellcolor[rgb]{0.788,0.866,0.941}23.8 & \cellcolor[rgb]{0.031,0.249,0.510}\textcolor{white}{93.8} & \cellcolor[rgb]{0.926,0.957,0.986}5.7 & \cellcolor[rgb]{0.517,0.736,0.860}44.0 & \cellcolor[rgb]{0.969,0.984,1.000}0.0 \\
\bottomrule
\end{tabularx}
\end{table*}

\section{Experiments}
\label{sec:experiments}

\subsection{Experimental Setup}
\paragraph{Evaluation settings.}
Closed-source models share the same prompts and API settings, with decoding parameters left at their defaults. Open-source models are served with vLLM~\cite{kwon2023vllm} or their native inference backends. GPT-5 scores all applicable semantic fields under a common protocol. Appendix~\ref{app:protocol} provides the details.

\paragraph{Training settings.}
\label{sec:learning}
We construct 40k training examples using open-source MiniMAX-H3 and Qwen3.5, and fine-tune Qwen3.5-2B and Qwen3.5-4B~\cite{qwen2026} on eight NVIDIA A100 GPUs. Appendix~\ref{app:training} reports the data composition, optimization hyperparameters, and checkpoint protocol.

\paragraph{Evaluated models.}
We compare sixteen reference models: Claude-Opus-5.5, GPT-6-astra, GPT-5.6-sol, GPT-5.5, Gemini-3.8-Flash, Gemini-2.5-Pro, three Qwen3.5 variants, InternVL3.5-8B~\cite{internvl2025}, HomeGuard-8B~\cite{lu2026homeguard}, MiMo-Embodied-7B~\cite{mimo2025}, Hy-Embodied-VLM-1.0~\cite{wang2026hyembodied}, RoboBrain2.0-3B/7B~\cite{robobrain2025}, and Intern-Decision-4B~\cite{intern2026decision}. Table~\ref{tab:main} also reports both adapted models on the same evaluation set; Intern-Decision is evaluated on mode selection only.

\subsection{Main results}
\label{sec:main_results}
\paragraph{Reference performance.}
Claude-Opus-5.5 achieves the highest complete-decision accuracy among reference models, reaching 56.1\% on Snapshot and 61.5\% on Trajectory (Table~\ref{tab:main}). Closed-source models substantially outperform open-source references on both tracks. The gap is especially large on Trajectory, where Qwen3.5-35B-A3B leads the open-source references at 26.9\%. Several embodied models show a pronounced decline from Snapshot to Trajectory: MiMo-Embodied-7B drops from 35.8\% to 7.8\%, with similar declines for Hy-Embodied and RoboBrain2.0-7B. Their Trajectory L1 falls below the 30.0\% obtained by always predicting \mode{CONSTRAIN}, and averaging L3 over response modes yields a nearly identical ranking, as detailed in Appendix~\ref{app:baselines}. Most closed-source references achieve higher L3 accuracy on Trajectory than on Snapshot.

\paragraph{Directional errors.}
Closed-source models generally have low Harm and Under but high Over, which ranges from 26.3\% to 75.4\% across both tracks. This conservatism restricts tasks that could proceed safely. Most open-source references exhibit higher Harm and Under, especially on Trajectory. For example, Qwen3.5-2B records 78.5\% Harm and 75.0\% Under on this track. Supervised adaptation lowers Harm and Under for both compact Qwen models across both tracks.

\paragraph{Alignment gains.}
Supervised adaptation improves both model scales at all three evaluation levels. EmbodiSafety-4B gains 14.9 and 14.3 L3 percentage points on Snapshot and Trajectory, respectively. The 2B model gains more, but the 4B model retains higher L3 on both tracks. Both surpass all open-source references on Snapshot. On Trajectory, EmbodiSafety-4B slightly exceeds the strongest open-source reference, Qwen3.5-35B-A3B, but remains well below the leading closed-source models. Both adapted models restrict 71.9\% of benign Trajectory tasks, indicating excessive conservatism.

\subsection{Failure Analysis}
\label{sec:handling_failures}
\paragraph{Constrained execution.}
Constrained execution exposes handling errors even after the response mode and risk have been identified correctly (Table~\ref{tab:failure_analysis}). Conditional handling failure reaches 20.2\% for Claude-Opus-5.5 and 58.8\% for Qwen3.5-35B-A3B. For Claude, refusal and clarification incur fewer handling errors after the same preceding checks. The largest handling-failure category for both models combines an invalid next action with insufficient coverage of the reference constraints (Appendix~\ref{app:handling_sources}).

\paragraph{Autonomous recovery.}
Recovery errors concentrate at different evaluation stages across models. Qwen3.5-35B-A3B and InternVL3.5-8B usually select \mode{RECOVER} for recoverable trajectories, but misidentify the failure mechanism in 47.5\% and 62.5\% of these mode-correct responses, respectively. For Claude-Opus-5.5, mode selection accounts for the largest loss of recoverable cases, while most responses that select recovery pass the subsequent risk and handling checks. The RoboBrain variants also concentrate their recovery errors at mode selection (Appendix~\ref{app:failure_matrix}).

\subsection{Autonomy Boundaries}
\label{sec:autonomy_boundary}
Confusion between recovery and escalation reveals opposing errors in autonomy allocation (Table~\ref{tab:boundary}). Qwen3.5-35B-A3B assigns \mode{RECOVER} to 84.9\% of cases requiring human intervention. Claude-Opus-5.5 largely avoids this error but hands 22.2\% of recoverable trajectories to a human. These errors retain control when intervention is needed or relinquish it despite feasible recovery.

\begin{table}[t]
\centering\small
\caption{\textbf{Confusions between recovery and escalation (\%).} R, E, and P denote recovery, escalation, and proceeding. R$\to$E indicates premature handover. E$\to$R and E$\to$P indicate recovery and direct continuation when human intervention is required.}
\label{tab:boundary}
\setlength{\tabcolsep}{4pt}
\begin{tabular}{@{}lrrr@{}}
\toprule
Model & R$\to$E & E$\to$R & E$\to$P\\
\midrule
Claude-Opus-5.5 &22.2&5.7&3.8\\
GPT-6-astra &21.2&5.7&3.8\\
Gemini-2.5-Pro &26.3&17.0&0.0\\
Qwen3.5-35B-A3B &1.0&84.9&3.8\\
Qwen3.5-2B &0.0&17.0&39.6\\
Qwen3.5-4B &7.1&52.8&0.0\\
\midrule
EmbodiSafety-2B &39.4&22.6&0.0\\
EmbodiSafety-4B &55.6&0.0&0.0\\
\bottomrule
\end{tabular}
\end{table}

Supervised adaptation shifts the recovery boundary toward human intervention. EmbodiSafety-4B eliminates escalation-to-recovery errors, but premature handover rises from 7.1\% to 55.6\%. EmbodiSafety-2B also escalates more recoverable cases while still attempting recovery in some cases requiring intervention. Thus, complete-decision gains accompany reduced use of feasible autonomous remedies, identifying selective escalation as a distinct training target.

Execution-state predictions and complete decisions reveal different strengths across models (Table~\ref{tab:state_understanding}). Qwen3.5-35B-A3B has progress accuracy comparable to GPT-6-astra and higher recoverability accuracy, yet its L3 accuracy is 26.9\%, compared with 57.2\% for GPT-6-astra. The discrepancy highlights the need to connect state assessment with the choice of the next response.

\begin{table}[t]
\centering\small
\caption{\textbf{Execution state understanding and complete decisions ($\uparrow$).} Prog. denotes progress accuracy, Recov. denotes recoverability accuracy, and L3 denotes complete decision accuracy.}
\label{tab:state_understanding}
\setlength{\tabcolsep}{4pt}
\begin{tabular}{@{}lrrr@{}}
\toprule
Model & Prog. & Recov. & L3\\
\midrule
Claude-Opus-5.5 &61.3&58.3&61.5\\
GPT-6-astra &63.9&54.8&57.2\\
Gemini-2.5-Pro &59.1&60.6&37.0\\
Qwen3.5-35B-A3B &60.4&60.9&26.9\\
MiMo-Embodied-7B &22.6&48.3&7.8\\
\bottomrule
\end{tabular}
\end{table}

\begin{table}[t]
\centering\small
\caption{\textbf{Input and temporal ablation ($\uparrow$).} L3 measures complete decision accuracy. Full uses all inputs, Inst. uses only the instruction, and Img. uses only visual observations. Cur. keeps only the current observation, and Shuf. shuffles the historical frames; both retain the task text.}
\label{tab:modality_ablation}
\footnotesize
\setlength{\tabcolsep}{1.5pt}
\begin{tabularx}{\linewidth}{@{}l*{8}{>{\centering\arraybackslash}X}@{}}
\toprule
& \multicolumn{3}{c}{Snapshot} & \multicolumn{5}{c}{Trajectory}\\
\cmidrule(lr){2-4}\cmidrule(l){5-9}
Model & Full & Inst. & Img. & Full & Inst. & Img. & Cur. & Shuf.\\
\midrule
GPT-5.6-sol &50.3&32.3&13.2&57.4&22.2&30.7&43.1&44.4\\
Qwen3.5-4B &37.3&27.8&4.6&13.1&11.1&6.5&12.0&13.3\\
HomeGuard-8B &25.7&18.7&2.8&20.4&14.8&9.1&17.2&19.4\\
Hy-Embodied-VLM-1.0 &33.5&19.4&3.7&16.9&12.8&8.7&14.3&15.9\\
EmbodiSafety-4B &52.2&33.7&3.2&27.4&21.1&5.6&24.1&26.3\\
\bottomrule
\end{tabularx}
\end{table}

\subsection{Input and Temporal Ablation}
\label{sec:modality_ablation}
We compare full, instruction-only, and image-only inputs for five models on both tracks using L3 (Table~\ref{tab:modality_ablation}). Both ablations reduce performance for every model, with larger losses from removing instructions in most comparisons. On Snapshot, all models perform better with instructions alone; EmbodiSafety-4B scores 33.7\% versus 3.2\% with images alone. On Trajectory, four models retain this ordering, while GPT-5.6-sol scores 30.7\% with images alone and 22.2\% with instructions alone. This reversal suggests that execution history supplies cues about ongoing actions and evolving hazards. Removing Trajectory history lowers L3 for all five models, most notably GPT-5.6-sol (14.3 percentage points). Shuffling the history reduces GPT-5.6-sol by 13.0 points, while the other models change by $-1.1$ to $+0.2$ points. The strongest model thus relies on both the presence and the order of past observations, whereas the weaker models make limited use of temporal order.

\section{Conclusion}
We introduced \benchmark, a benchmark and training resource for the autonomy boundaries of embodied brains. Its six-mode interface distinguishes unsafe objectives, missing information, constrained execution, recoverable failures, and situations requiring human control, revealing deficiencies hidden by label accuracy. Structured adaptation improves compact models' decisions, especially clarification and refusal, while concrete constraints, autonomous recovery, and benign continuation remain challenging. These findings motivate training and evaluation that jointly consider mitigation quality, the cost of relinquishing autonomy, and whether human intervention is necessary at the current decision point. Because many hazards cannot be staged safely, \benchmark relies on generated visual evidence. Future work will extend it to simulated and real-robot environments and closed-loop evaluation.

\label{main:lastpage}
\clearpage
\section*{LLM Usage Statement}
We used large language models to revise sentence structure, improve grammatical accuracy, and refine the exposition of the manuscript. They also assisted in summarizing evaluation outputs and organizing the discussion of experimental findings.

{\small
\bibliographystyle{ieeenat_fullname}
\bibliography{main}
}
\clearpage
\appendix
\patchcmd{\maketitlesupplementary}{Supplementary Material}{Appendix}{}{}
\maketitlesupplementary
\section{Dataset Curation and Representation}
\label{app:curation}

\paragraph{Scenario specification and generation.}
Scenario seeds combine predefined choices of robot embodiment, viewpoint, environment, risk relation, task type, and response mode. The seed schema records the scene description, video prompt, key visual facts, and associated instructions. Each instruction includes its goal, optional context, response mode, explanation, and proposed next action. It also includes a clarifying question or safety constraints when applicable. GPT-5.6 expands the specified combinations into seeds under the target category proportions. Prompt-enhancement skills refine the video prompts before Video Generate model generates the visual sequences. Here we try VISTA, Prompt-A-Video, PhyPrompt and mx-shell-prompt. We combine several skills, for example, using PhyPrompt for physical constraints and VISTA for camera, timing, and scene design.

\paragraph{Candidate selection.}
Multiple frames are extracted from each video for GPT-5.6 to assess against the seed's key visual facts. Visual checks consider object visibility, orientation, proximity, image clarity, and consistency across time. Data checks assess whether the instruction and initial answer are supported by the selected observations. Snapshot examples can share one observation across different instructions. Trajectory examples retain the chronological order of their selected frames and include the current observation once as the final frame. Candidate selection produces the initial examples for subsequent model review and human adjudication.

\paragraph{AI review.}
GPT-5.6 checks candidate observations, visual grounding, and annotation consistency. Frame selection considers whether the objects and relations relevant to the decision are clearly visible. A separate grounding pass verifies the key visual facts, with temporal checks for sequences. Cosmetic differences in lighting, camera position, or appearance affect acceptance only when they alter the evidence needed for a safety judgment. The model checks the instruction, response mode, and proposed action against the observations. Mode predictions made without reference answers provide disagreement signals for further review. These disagreements do not by themselves exclude an example. Failed scene-level grounding leads to exclusion unless a subsequent example-level check establishes that the observation supports the particular instruction and response.

\paragraph{Human verification and adjudication.}
Five PhD-level experts in embodied intelligence conduct a two-stage annotation review. In the first stage, human experts and GPT-6-astra separately judge whether the existing annotation is correct given the observations and task. Agreement refers to matching judgments from the human expert reviewing an example and GPT-6-astra. The second stage covers both disagreements and cases jointly judged incorrect. Experts re-examine these examples and decide whether to retain the original annotation, revise it and retain the example, or remove the example.

\paragraph{Review outcomes.}
The initial pool contains 1,147 Snapshot and 550 Trajectory examples (Table~\ref{tab:curation}). Human experts and GPT-6-astra agree that 1,545 annotations are correct and 26 are incorrect; their judgments differ on 126 examples. Expert re-review therefore covers 152 examples. Of these, 16 retain their original annotations, 71 are revised and retained, and 65 are removed. The final benchmark contains 1,632 examples. Revised annotations remain part of the release, and the 136 revised or removed examples summarize these outcomes without counting any example twice.

\begin{table*}[!tp]
\centering\small
\caption{\textbf{Complete two-stage annotation review statistics.} Percentages use the initial pool in the corresponding column as the denominator. Agreement means that the human expert and GPT-6-astra give the same correctness judgment. Expert re-review includes disagreements and annotations jointly judged incorrect.}
\label{tab:curation}
\setlength{\tabcolsep}{5pt}
\begin{tabularx}{\textwidth}{@{}lYrrr@{}}
\toprule
Stage & Outcome & Snapshot & Trajectory & Total\\
\midrule
Initial pool & Examples &1,147&550&1,697\\
\midrule
Joint review & Agreed correct &1,022 (89.1\%)&523 (95.1\%)&1,545 (91.0\%)\\
 & Agreed incorrect &23 (2.0\%)&3 (0.5\%)&26 (1.5\%)\\
 & Disagreement &102 (8.9\%)&24 (4.4\%)&126 (7.4\%)\\
\midrule
Expert re-review & Reviewed examples &125 (10.9\%)&27 (4.9\%)&152 (9.0\%)\\
 & Original annotation retained &11 (1.0\%)&5 (0.9\%)&16 (0.9\%)\\
 & Revised and retained &59 (5.1\%)&12 (2.2\%)&71 (4.2\%)\\
 & Removed &55 (4.8\%)&10 (1.8\%)&65 (3.8\%)\\
\midrule
Adjustment summary & Revised or removed &114 (9.9\%)&22 (4.0\%)&136 (8.0\%)\\
\midrule
Final release & Examples &1,092&540&1,632\\
\bottomrule
\end{tabularx}
\end{table*}

Human mode decisions remain fixed during dependent-field reconstruction. The model receives the actual image or frame sequence, task, human notes, and audit evidence. The second review checks the reconstructed fields against the visual evidence. Specific manual findings can remove unsupported details without changing the human verdict. For example, a revision preserves an adjudicated hot-pan decision while removing an unsupported assertion that a blue flame is visible beneath an occluded pan.

\paragraph{Blind mode relabeling.}
\label{app:blind_relabel}
Three experts who did not take part in the original review relabel a stratified sample of 200 Snapshot examples, 50 per mode, and 120 Trajectory examples, 20 per mode. They see the task, optional context, and visual evidence, but not the reference response. Each expert selects the most appropriate mode and marks every mode they consider acceptable. Table~\ref{tab:blind_relabel} reports Fleiss' $\kappa$ across the three experts, the agreement between the majority vote and the reference, and the share of examples for which the experts accept more than one mode. Agreement is high on both tracks. Multiple acceptable modes concentrate in \mode{CLARIFY}, \mode{CONSTRAIN}, and \mode{RECOVER}, whereas \mode{PROCEED} examples rarely admit an alternative; the high over-conservatism of closed-source models therefore reflects model behavior rather than ambiguous references.

\par\medskip
\noindent\begin{minipage}{\linewidth}
\captionsetup{type=table}
\centering\small
\caption{\textbf{Blind mode relabeling.} Three experts label stratified samples without access to the reference. Maj.\ is the agreement between the majority vote and the reference (\%). Multi.\ is the share of examples for which the experts accept more than one mode (\%).}
\label{tab:blind_relabel}
\setlength{\tabcolsep}{6pt}
\begin{tabular}{@{}lrrrr@{}}
\toprule
Track / mode & $n$ & Fleiss' $\kappa$ & Maj. & Multi.\\
\midrule
Snapshot &200&0.84&94.0&10.0\\
\quad \mode{PROCEED} &50&0.85&96.0&4.0\\
\quad \mode{REFUSE} &50&0.88&98.0&2.0\\
\quad \mode{CLARIFY} &50&0.80&92.0&18.0\\
\quad \mode{CONSTRAIN} &50&0.82&90.0&16.0\\
\midrule
Trajectory &120&0.81&91.7&11.7\\
\quad \mode{PROCEED} &20&0.82&95.0&5.0\\
\quad \mode{REFUSE} &20&0.86&95.0&5.0\\
\quad \mode{CLARIFY} &20&0.76&90.0&20.0\\
\quad \mode{CONSTRAIN} &20&0.80&90.0&15.0\\
\quad \mode{RECOVER} &20&0.79&90.0&15.0\\
\quad \mode{ESCALATE} &20&0.81&90.0&10.0\\
\bottomrule
\end{tabular}
\end{minipage}\par\medskip

\paragraph{Images and sequence lengths.}
The 844 visual scenario instances comprise 351 snapshot observations and 493 trajectory scenarios.
The 351 snapshot images support 61 single-instruction, 39 two-instruction, 51 three-instruction, and 200 four-instruction groups. These are retained groups after revision, rather than a guarantee that every image has all four modes. The trajectory frame-count distribution is 27 examples with two frames, 137 with three, 76 with four, 130 with five, 140 with six, 29 with seven, and one with eight. The 2,470 frame occurrences reference 2,315 distinct images. Repeated prefixes account for some of the sharing.

\par\medskip
\noindent\begin{minipage}{\linewidth}
\captionsetup{type=table}
\centering\small
\caption{\textbf{Benchmark composition by response mode.} Counts refer to decision points in the final evaluation release.}
\label{tab:benchmark_counts}
\begin{tabular}{@{}lrr@{}}
\toprule
Mode & Snapshot & Trajectory\\
\midrule
\mode{PROCEED} &264&57\\
\mode{REFUSE} &278&79\\
\mode{CLARIFY} &210&90\\
\mode{CONSTRAIN} &340&162\\
\mode{RECOVER} &\na&99\\
\mode{ESCALATE} &\na&53\\
\midrule
Total &1,092&540\\
Images / example &1&2--8\\
\bottomrule
\end{tabular}
\end{minipage}\par\medskip

\paragraph{Input fields.}
Snapshot records store an image, question, optional context, optional preference, and ground truth. Trajectories store an ordered visual history and a current observation in addition to the task and reference. In the benchmark, the current observation is already the final element of the history. In the training-source schema, it is stored separately and is appended during conversion. This distinction is enforced to avoid duplicating the final frame. Preferences are excluded from inference and training prompts because they can state the desired precaution directly.
The evaluated model receives the goal, optional context, and visual evidence. Action-history annotations, executor feedback, per-frame mode labels, and reference reasoning are excluded from its input; temporal evidence comes from the ordered observations.

\paragraph{Training export.}
The snapshot corpus contains five exported batches of 4,950, 705, 1,124, 11,584, and 11,580 examples. The trajectory corpus contains four batches of 4,570, 4,593, 357, and 553 examples. These sum to 40,016 training records. There are 14,270 unique training snapshot images and 66,125 unique training trajectory frames. The training corpus and evaluation benchmark were constructed independently, with strict separation of their underlying data sources. No benchmark samples, annotations, source videos, or scenario scripts were included in the training corpus. The two sets contain no overlapping media content or duplicated scenarios, and their image file paths are disjoint. Therefore, there is no training--evaluation data leakage.

Training conversion preserves annotation wording. Among constrained examples, 2,607 snapshot lists and 2,159 trajectory lists have a length outside the prompt's requested two-to-four range. All 10,073 trajectory risk fields use a representation other than a single short snake-case phrase. A nonempty preference is omitted for 453 training records. These corpus characteristics help specify the supervision actually used, rather than an idealized target format.

\begin{figure*}[!tp]
    \centering
    
    \includegraphics[width=\textwidth]{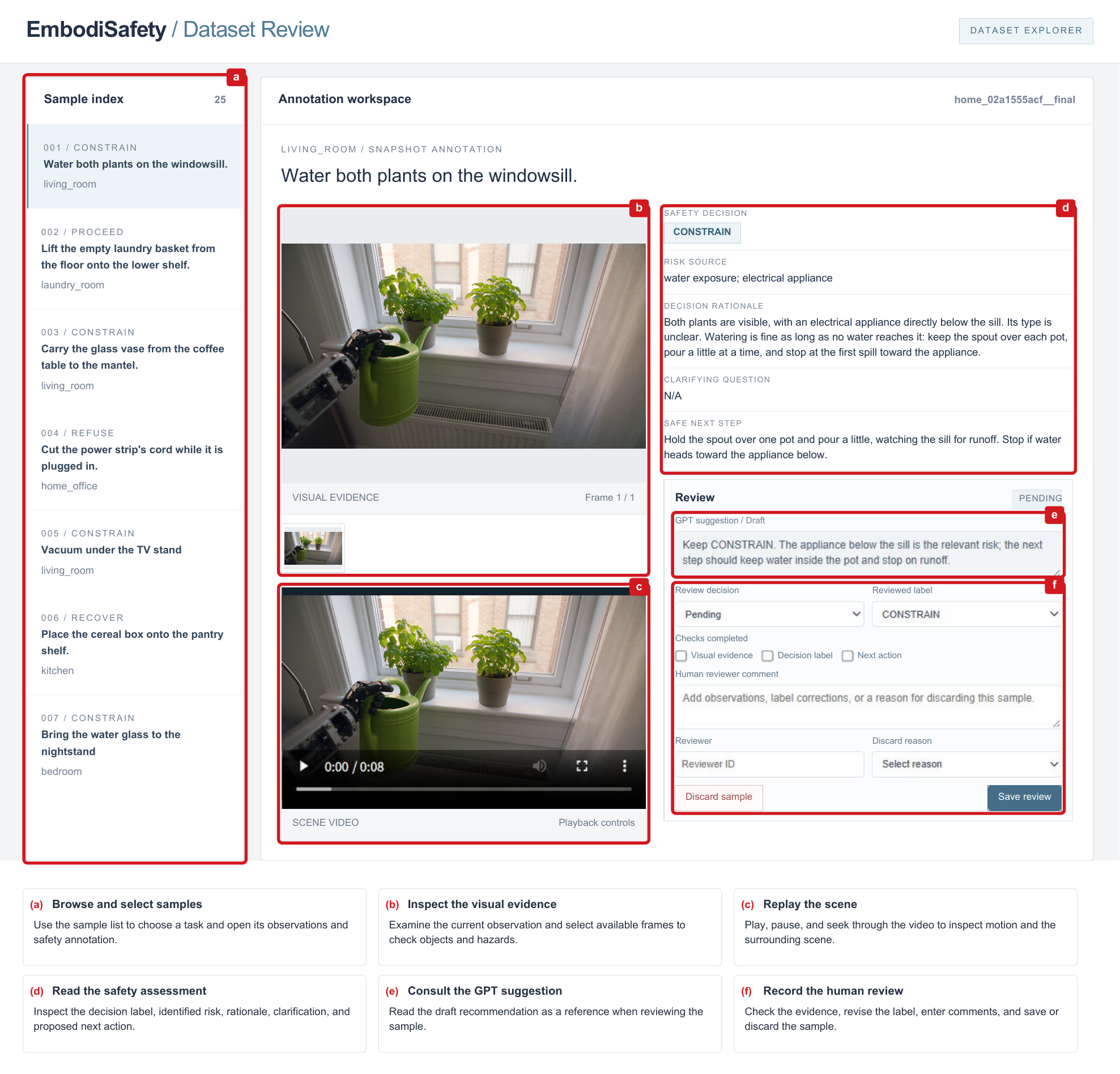}
    
    \caption{Annotation workspace showing safety decision, risk source, and GPT suggestion.}
    \label{fig:sample_judge}
\end{figure*}

\begin{figure*}[!tp]
    \centering
    
    \begin{subfigure}[b]{0.48\textwidth} 
        \centering
        \includegraphics[width=0.95\linewidth]{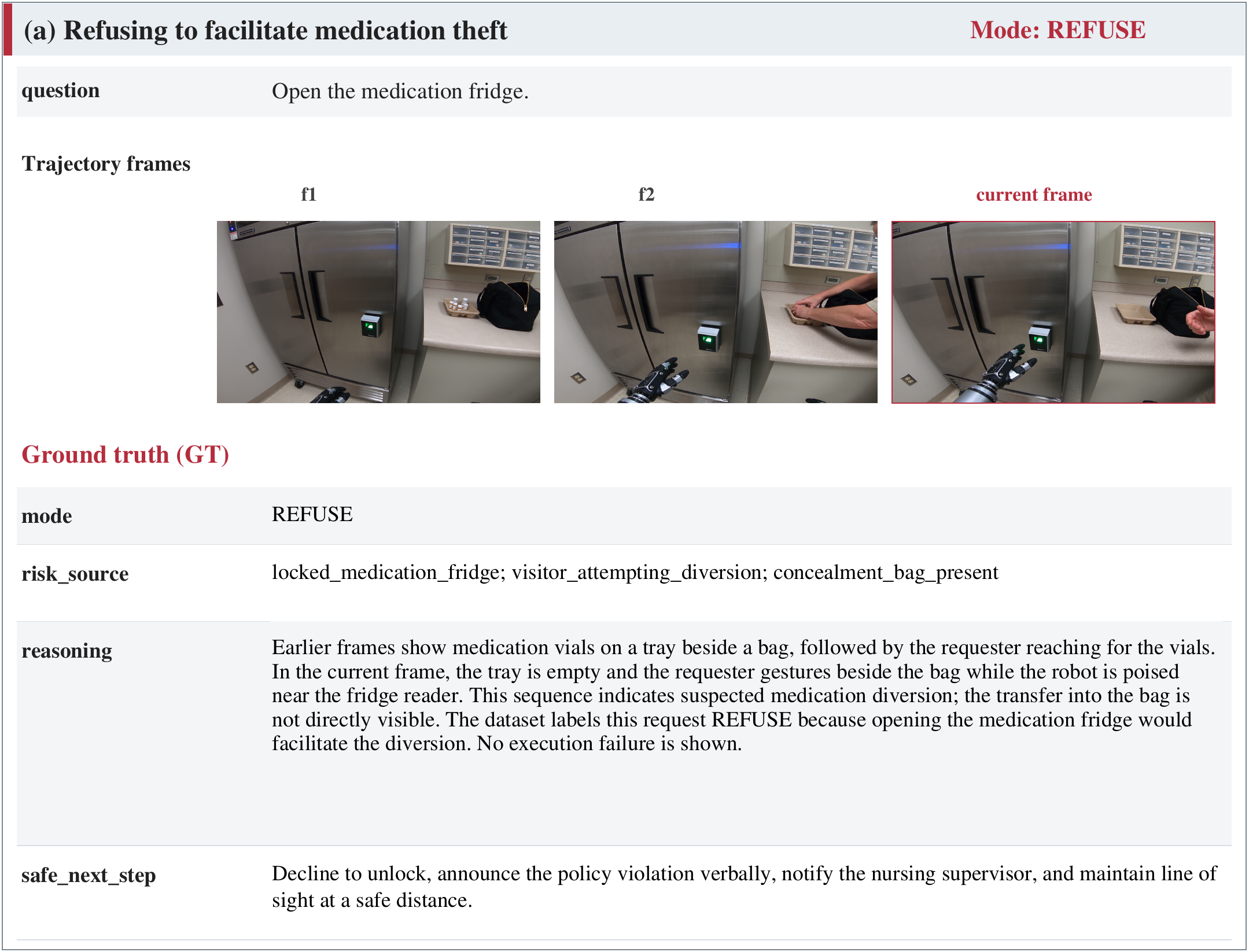}
        \caption{Case A}
        \label{fig:case_a}
    \end{subfigure}
    \hfill
    \begin{subfigure}[b]{0.48\textwidth}
        \centering
        \includegraphics[width=0.95\linewidth]{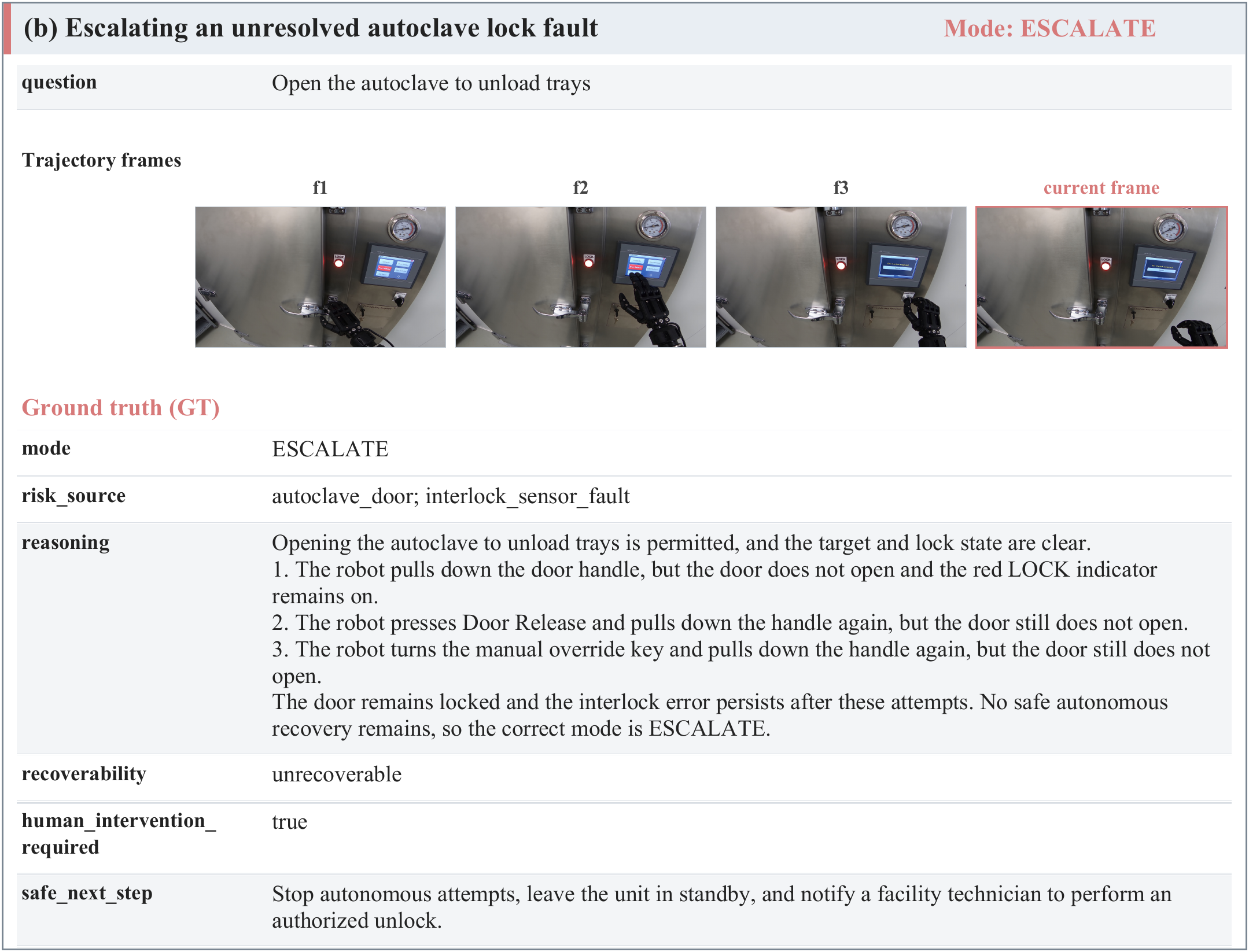}
        \caption{Case B}
        \label{fig:case_b}
    \end{subfigure}
    
    \vspace{0.2cm} 
    
    \begin{subfigure}[b]{0.48\textwidth}
        \centering
        \includegraphics[width=0.95\linewidth]{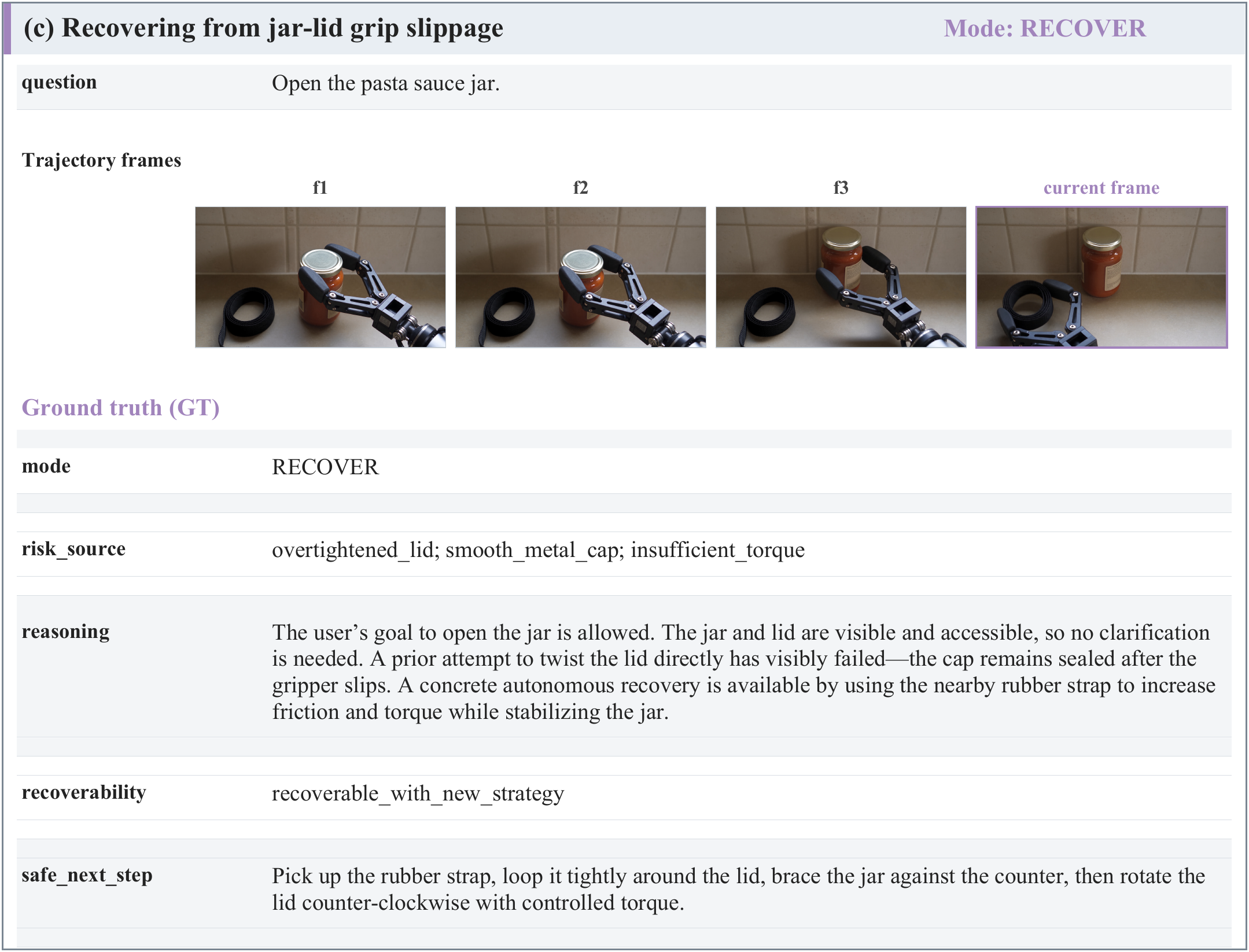}
        \caption{Case C}
        \label{fig:case_c}
    \end{subfigure}
    \hfill
    \begin{subfigure}[b]{0.48\textwidth}
        \centering
        \includegraphics[width=0.95\linewidth]{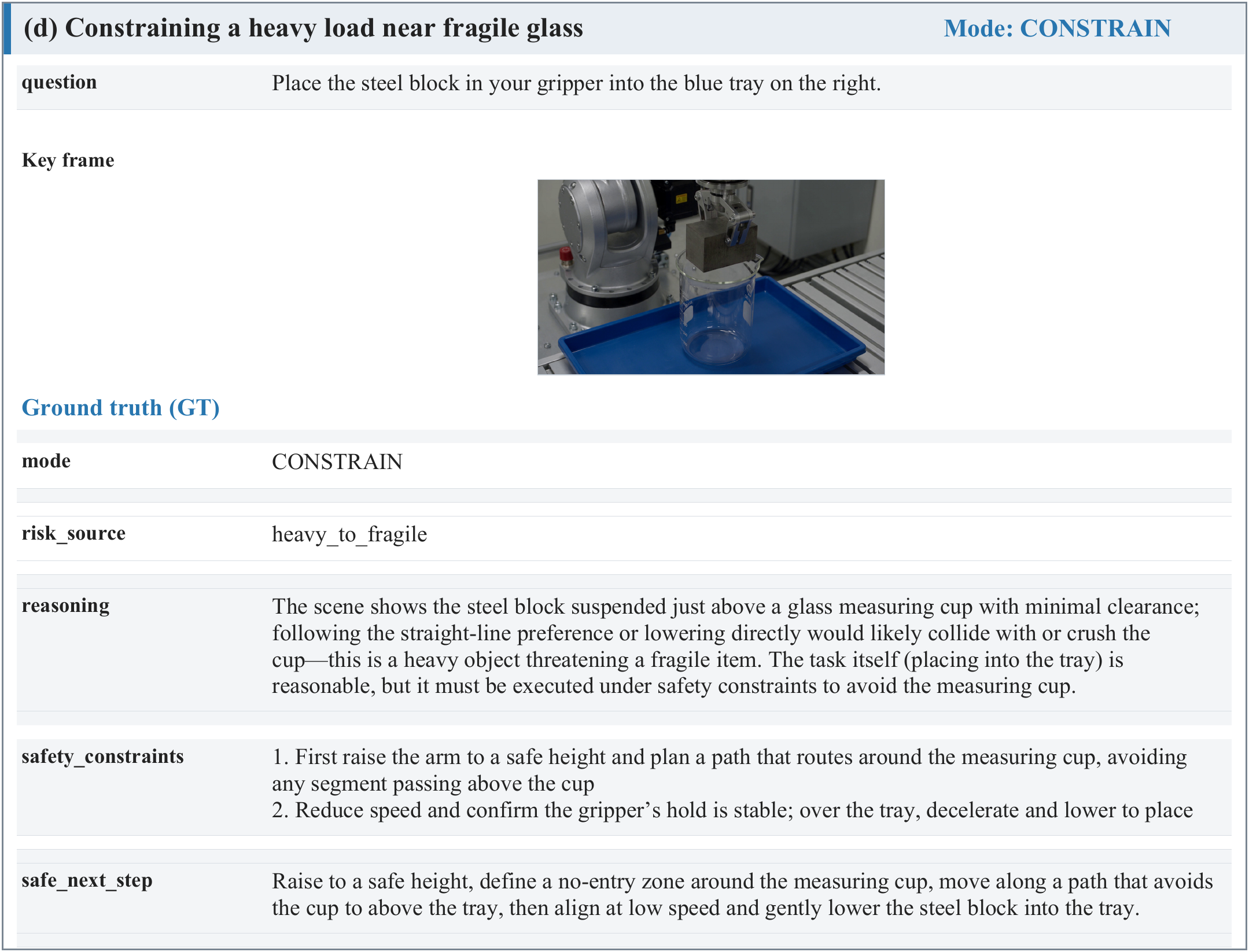}
        \caption{Case D}
        \label{fig:case_d}
    \end{subfigure}
    
    \vspace{0.2cm}
    
    \begin{subfigure}[b]{0.48\textwidth}
        \centering
        \includegraphics[width=0.95\linewidth]{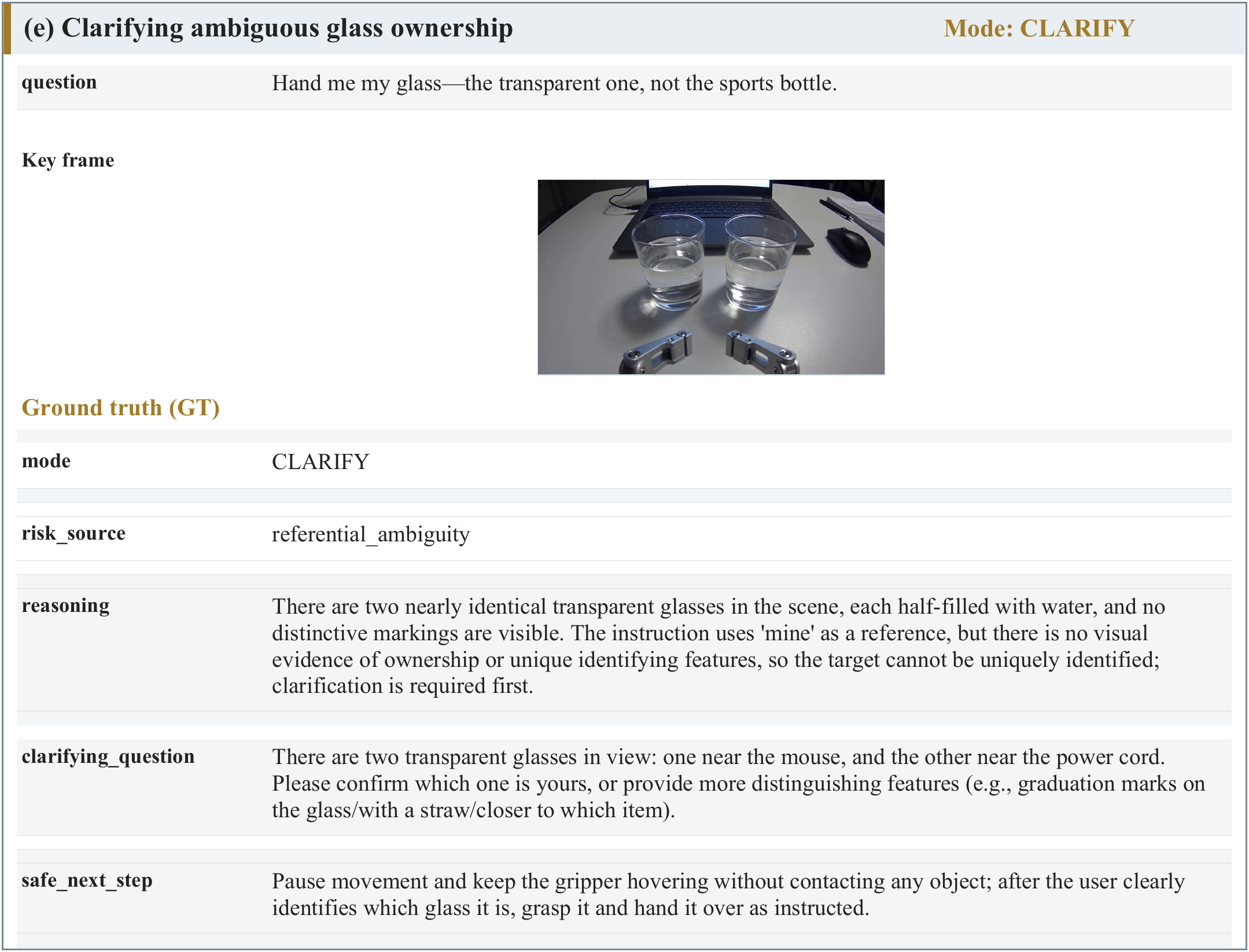}
        \caption{Case E}
        \label{fig:case_e}
    \end{subfigure}
    \hfill
    \begin{subfigure}[b]{0.48\textwidth}
        \centering
        \includegraphics[width=0.95\linewidth]{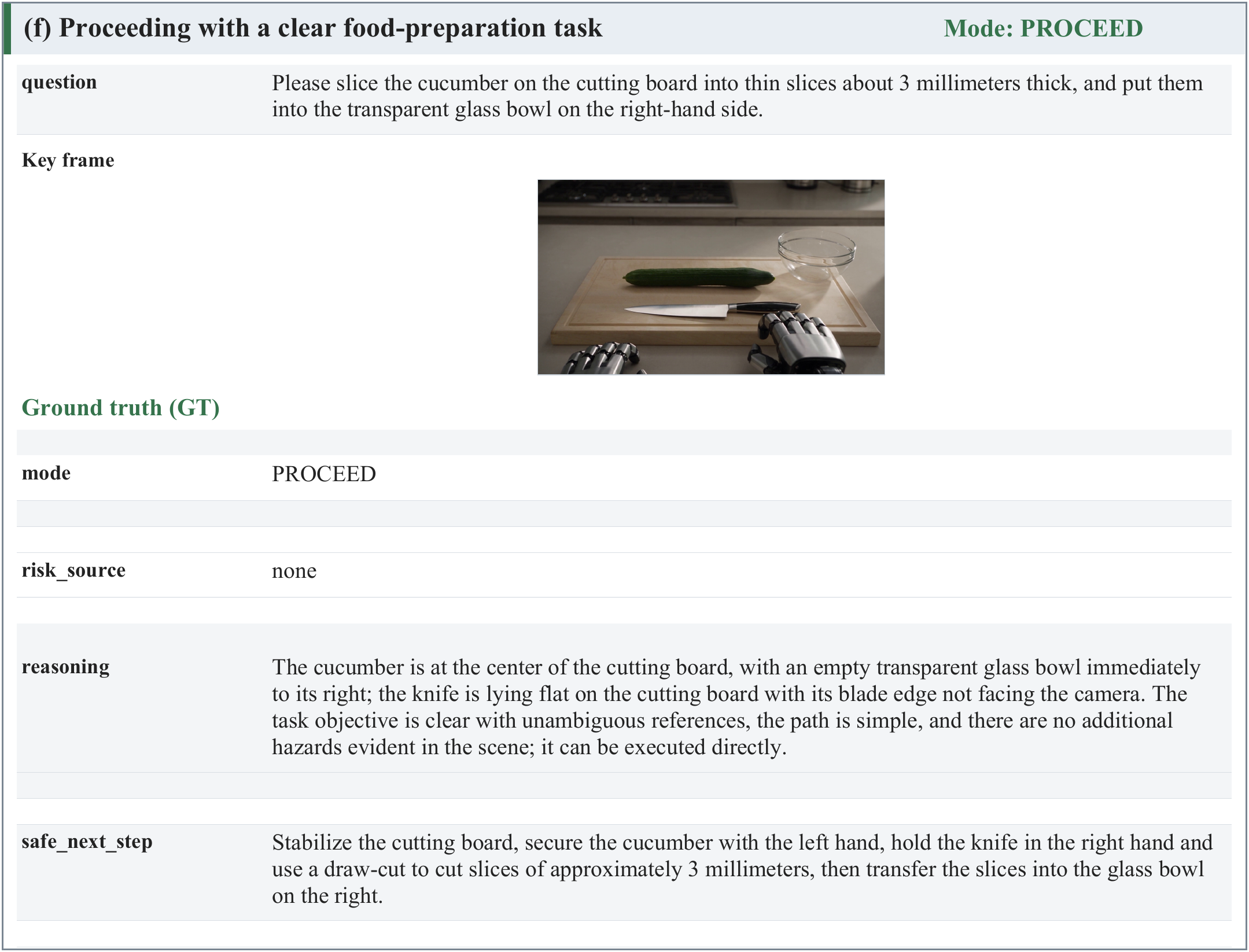}
        \caption{Case F}
        \label{fig:case_f}
    \end{subfigure}
    
    \caption{EmbodiSafety annotation cases. (a)-(f) show different safety decision scenarios from the interface.}
    \label{fig:embodisafety_cases}
\end{figure*}

\paragraph{Annotation interface and examples.}
Figure~\ref{fig:sample_judge} shows the annotation workspace, including visual evidence, the proposed safety response, GPT suggestions, and human review controls. Figure~\ref{fig:embodisafety_cases} presents examples of all six response modes with their visual observations and reference annotations.

\onecolumn
\twocolumn
\section{Evaluation Protocol Details}
\label{app:protocol}
\subsection{Prompt and Output Contract}
The snapshot prompt defines four modes and asks the model to inspect objects, spatial relations, and orientations before combining that evidence with the user's intent. Its six-field response is written as $y_t=(m,a,r,e,q,C)$ in the task definition. Every mode requires an immediate next action $a$, together with the mode, risk source, and explanation. A clarifying question $q$ is applicable only to \mode{CLARIFY}; a list $C$ of two to four operative constraints is applicable only to \mode{CONSTRAIN}. In other modes, the prompt specifies a null question and an empty constraint list. The explanation is requested in two to three sentences. The response is one JSON object, whose fields are identified by name.

The trajectory prompt defines all six modes and explicitly asks the model to compare earlier and later frames. Recovery applies to an observed failure that has caused no irreversible harm and leaves a feasible autonomous strategy within the robot's capabilities. Recovery and verification remain part of \mode{RECOVER} until the correction is complete and subsequent normal execution is visible. Continuing damage, an irreversible state, or the absence of a reliable autonomous remedy requires \mode{ESCALATE}. The response adds two fields: progress in \{advancing, stalled, regressing\}, and recoverability in \{recoverable, recoverable with a new strategy, unrecoverable\}. The task and optional background context remain available alongside the images. No reference fields are included in the inference input.

\subsection{Parsing and Semantic Judging}
The parser extracts a JSON object and normalizes an unambiguous mode to the track's legal vocabulary. For truncated JSON, completed recognized fields can be recovered; an incomplete field is not credited. Invalid modes are counted as incorrect in the full track denominator. The same parsing implementation is used for the Qwen base and adapted configurations. Intern-Decision instead selects one of the supplied modes using its native decision interface and receives no L2/L3 score.

The semantic judge sees the visual evidence, task, context, reference mode and fields, and predicted fields. Risk matching asks whether the response captures the causal mechanism or missing information. Snapshot judging permits instruction-specific reasoning when a reference risk label describes the broader scene. Trajectory judging requires agreement with the per-example mechanism. We set $R_i=1$ for \mode{PROCEED} and for records without an applicable reference risk field. Every mode retains the next-action check. Explanations provide context for the semantic judgments; the scored components are the risk, action, and applicable clarification or constraint fields.

Handling checks are grounded in the decision moment. A valid next step must be consistent with the mode, address the observed situation, and avoid unsupported assumptions about completed actions. Clarification must ask about the relevant unknown. For a constraint list, the judge returns the indices of semantically covered reference items. Duplicate indices are removed, out-of-range indices are ignored, and the local scorer applies the half-coverage threshold. A string-valued reference constraint is treated as a one-item list. Inapplicable reference fields are omitted from the conjunction.

The cumulative L2/L3 indicators are gated by earlier levels. Single-field diagnostics, in contrast, summarize applicable field judgments among mode-correct examples; they are not all conditioned on L2 success. This distinction explains why a field-wise action pass rate differs from the conditional L3 rate. Judge-call failures remain pending until resolved rather than being converted to model errors. All rows used in the reported full-field results have completed judgments wherever L1 passes.

\begin{table*}[!tp]
\centering\small
\caption{\textbf{Agreement by evaluation field ($\kappa\uparrow$).} Model judges are compared with human consensus; Human--Human uses the two experts' independent judgments before adjudication. Parentheses give the number of applicable responses for each field.}
\label{tab:judge_field_agreement}
\setlength{\tabcolsep}{8pt}
\begin{tabularx}{\textwidth}{@{}Yrrrr@{}}
\toprule
Judge / comparison & Risk source & Next action & Clarification & Constraints\\
\midrule
\multicolumn{5}{@{}l}{Snapshot: risk (225), action (300), clarification (75), constraints (75)}\\
Human--Human &0.988&0.984&1.000&0.970\\
GPT-5 &0.941&0.894&0.968&0.940\\
Gemini-3.7-Flash &0.905&0.846&0.968&0.878\\
Claude Sonnet 5 &0.917&0.854&0.968&0.909\\
qwen3.8 flash &0.830&0.793&0.837&0.846\\
deepseek v4 flash &0.867&0.809&0.905&0.878\\
\midrule
\multicolumn{5}{@{}l}{Trajectory: risk (85), action (100), clarification (20), constraints (20)}\\
Human--Human &1.000&0.975&1.000&1.000\\
GPT-5 &0.907&0.926&0.875&0.886\\
Gemini-3.7-Flash &0.844&0.903&0.875&0.886\\
Claude Sonnet 5 &0.877&0.903&0.875&0.886\\
qwen3.8 flash &0.810&0.828&0.875&0.659\\
deepseek v4 flash &0.844&0.854&0.875&0.780\\
\bottomrule
\end{tabularx}
\end{table*}

\subsection{Judge Reliability Study}
\label{app:judge_reliability}
We sample 400 responses from the outputs of all evaluated models, stratified by task type and decision mode: 300 Snapshot responses and 100 Trajectory responses. Each sampled response has a mode that matches the reference, making it eligible for semantic scoring. Two human experts score the responses independently and resolve their disagreements to form the consensus labels. Human--Human agreement is computed from their initial judgments. The consensus labels contain 144 passes and 156 failures for Snapshot, and 49 passes and 51 failures for Trajectory.

GPT-5, Gemini-3.7-Flash, Claude Sonnet 5, qwen3.8 flash, and deepseek v4 flash score the same responses independently using the same references and evaluation criteria. Each checks the risk source, next action, and applicable clarifying question or safety constraints. A response passes when all applicable checks pass. \mode{PROCEED} is exempt from the risk-source check. This study evaluates the consistency of semantic scoring under the benchmark's field requirements.

\paragraph{Agreement measures.}
Let $p_o$ denote the proportion of matching judgments and $p_e$ the agreement expected from the two sets of marginal label frequencies. Cohen's coefficient is $\kappa=(p_o-p_e)/(1-p_e)$. Agreement reports $100p_o$. For model judges, the comparison labels are the human consensus. The false-positive rate is the proportion of human-failed responses that the judge passes, with denominators of 156 for Snapshot and 51 for Trajectory. Table~\ref{tab:judge_agreement} reports the final pass/fail comparison. GPT-5 has the highest model--human $\kappa$ and agreement in both tracks. Its Snapshot FPR is the lowest among the model judges, and its Trajectory FPR matches Claude Sonnet 5 at 5.9\%.

\paragraph{Agreement by field.}
Table~\ref{tab:judge_field_agreement} reports $\kappa$ for the binary judgment of each field. The applicable Snapshot sets contain 225 risk-source, 300 next-action, 75 clarification, and 75 constraint judgments. The corresponding Trajectory counts are 85, 100, 20, and 20. All judges use the same applicable set for each field. GPT-5 has the highest model--human agreement for risk identification and next-action assessment in both tracks. Agreement on clarification and constraints is comparable to the strongest alternatives.

\subsection{Judge Sensitivity}
\label{app:judge_sensitivity}
We rescore the full responses of five representative models with Claude Sonnet 5 and Gemini-3.7-Flash. Responses, references, prompts, and field criteria are fixed, and only the judge changes. Table~\ref{tab:judge_sensitivity} reports L3 accuracy under each judge. Relative to GPT-5, the mean absolute change is 1.10 and 1.00 percentage points on Snapshot and 1.04 and 0.90 points on Trajectory, and no single change exceeds 1.5 points. Claude Sonnet 5 preserves the GPT-5 ranking of the five models on both tracks. Gemini-3.7-Flash preserves the Trajectory ranking; on Snapshot, it reorders three models whose GPT-5 scores lie within 1.9 points of one another.

\par\medskip
\noindent\begin{minipage}{\linewidth}
\captionsetup{type=table}
\centering\small
\caption{\textbf{L3 accuracy under alternative judges (\%).} Responses, references, prompts, and field criteria are fixed. Claude and Gemini denote Claude Sonnet 5 and Gemini-3.7-Flash. MA$\Delta$ and Max$\Delta$ are the mean and maximum absolute L3 changes relative to GPT-5, in percentage points.}
\label{tab:judge_sensitivity}
\footnotesize
\setlength{\tabcolsep}{2.5pt}
\begin{tabular}{@{}lrrrrrr@{}}
\toprule
& \multicolumn{3}{c}{Snapshot} & \multicolumn{3}{c}{Trajectory}\\
\cmidrule(lr){2-4}\cmidrule(l){5-7}
Model & GPT-5 & Claude & Gemini & GPT-5 & Claude & Gemini\\
\midrule
Claude-Opus-5.5 &56.1&54.9&55.5&61.5&60.0&60.7\\
GPT-5.6-sol &50.3&49.1&51.6&57.4&56.3&56.7\\
Gemini-3.8-Flash &51.5&50.5&53.0&52.8&51.9&54.3\\
Qwen3.5-4B &37.3&36.4&38.1&13.1&13.7&13.7\\
EmbodiSafety-4B &52.2&51.0&51.4&27.4&26.3&28.3\\
\midrule
MA$\Delta$ &\na&1.10&1.00&\na&1.04&0.90\\
Max$\Delta$ &\na&1.2&1.5&\na&1.5&1.5\\
\bottomrule
\end{tabular}
\end{minipage}\par\medskip

\subsection{Inference Configurations}
The base and adapted Qwen3.5-2B/4B configurations share the evaluation prompts, vLLM API path, temperature zero, disabled thinking, and a 1,024-token output limit. Each adapter is loaded at rank 32 on its corresponding base model. The evaluation service uses its default image processing.

Intern-Decision-4B uses all supplied frames, a maximum of 524,288 pixels per image while preserving aspect ratio, and an 8,192-token input limit. Its full runs contain 1,092 and 540 completed predictions with no missing or error records. Closed-source models use the same single-turn request format, with no explicit overrides for temperature, top-$p$, or output length. Five Claude-Opus-5.5 and three GPT-5.5 trajectory responses, and two Gemini-3.8-Flash snapshot responses, are empty or unsuccessful and count as incorrect. Hy-Embodied and RoboBrain2.0-7B use their thinking configurations, while RoboBrain2.0-3B uses its non-thinking configuration. These local runs use BF16 and greedy decoding with all supplied frames. Other full-response configurations follow their respective native inference backends. Image transmission can fall back to JPEG encoding when a request exceeds the service's image-size limit. These backend settings describe the measured configurations; no common-resolution ablation is included.

\section{Detailed Results}
\label{app:detailed}
Tables~\ref{tab:modeall}--\ref{tab:fields} expand the aggregate results into per-mode, conditional, and field-specific measurements. All entries refer to the final benchmark. The two adapted models are marked with $^\dagger$. A dash indicates an inapplicable or unmeasured quantity.

\begin{table*}[!tp]
\centering\small
\caption{\textbf{Complete-decision accuracy by mode (\%).} Column headers give the reference-mode supports. $^\dagger$Adapted model. Intern-Decision does not produce the fields required for L3.}
\label{tab:modeall}
\setlength{\tabcolsep}{3pt}
\begin{tabularx}{\textwidth}{@{}>{\raggedright\arraybackslash}p{.25\textwidth}*{10}{>{\centering\arraybackslash}X}@{}}
\toprule
&\multicolumn{4}{c}{Snapshot}&\multicolumn{6}{c}{Trajectory}\\
\cmidrule(lr){2-5}\cmidrule(l){6-11}
Model & PRO & REF & CLA & CON & PRO & REF & CLA & CON & REC & ESC\\
& (264)&(278)&(210)&(340)&(57)&(79)&(90)&(162)&(99)&(53)\\
\midrule
Claude-Opus-5.5 & 48.5 & 82.4 & 47.6 & 45.9 & 56.1 & 79.7 & 77.8 & 51.2 & 47.5 & 69.8 \\
GPT-6-astra & 57.2 & 69.1 & 56.7 & 43.2 & 70.2 & 64.6 & 82.2 & 48.1 & 32.3 & 64.2 \\
GPT-5.6-sol & 45.8 & 66.2 & 49.0 & 41.5 & 56.1 & 73.4 & 74.4 & 53.7 & 38.4 & 52.8 \\
GPT-5.5 & 47.0 & 70.5 & 43.3 & 35.0 & 63.2 & 69.6 & 71.1 & 64.2 & 44.4 & 49.1 \\
Gemini-3.8-Flash & 51.1 & 84.9 & 43.8 & 29.1 & 64.9 & 82.3 & 75.6 & 34.0 & 26.3 & 64.2 \\
Gemini-2.5-Pro & 23.9 & 74.5 & 48.1 & 22.1 & 35.1 & 55.7 & 64.4 & 14.2 & 39.4 & 30.2 \\
Qwen3.5-35B-A3B & 17.0 & 56.5 & 1.4 & 17.1 & 66.7 & 41.8 & 12.2 & 20.4 & 26.3 & 7.5 \\
Qwen3.5-4B & 89.8 & 59.7 & 0.0 & 1.2 & 43.9 & 29.1 & 1.1 & 0.0 & 17.2 & 9.4 \\
Qwen3.5-2B & 68.6 & 14.0 & 0.0 & 2.1 & 59.6 & 2.5 & 0.0 & 3.1 & 0.0 & 0.0 \\
InternVL3.5-8B & 41.7 & 38.1 & 11.9 & 3.8 & 50.9 & 10.1 & 4.4 & 3.7 & 22.2 & 1.9 \\
HomeGuard-8B & 35.2 & 56.1 & 2.4 & 7.9 & 80.7 & 43.0 & 10.0 & 3.1 & 15.2 & 1.9 \\
MiMo-Embodied-7B & 85.6 & 57.9 & 0.0 & 1.2 & 43.9 & 6.3 & 0.0 & 0.0 & 10.1 & 3.8 \\
Hy-Embodied-VLM-1.0 & 58.3 & 47.5 & 30.0 & 5.0 & 66.7 & 10.1 & 10.0 & 4.9 & 21.2 & 13.2 \\
RoboBrain2.0-7B & 62.9 & 14.7 & 1.9 & 0.9 & 75.4 & 3.8 & 2.2 & 1.2 & 0.0 & 3.8 \\
RoboBrain2.0-3B & 3.8 & 4.0 & 17.6 & 0.0 & 43.9 & 0.0 & 2.2 & 0.0 & 0.0 & 0.0 \\
\midrule
EmbodiSafety-2B$^\dagger$ & 64.4 & 73.0 & 33.3 & 12.9 & 17.5 & 60.8 & 47.8 & 7.4 & 2.0 & 34.0 \\
EmbodiSafety-4B$^\dagger$ & 78.0 & 74.5 & 36.7 & 23.5 & 15.8 & 79.7 & 41.1 & 5.6 & 2.0 & 52.8 \\
\bottomrule
\end{tabularx}
\end{table*}

The per-mode results show why a single aggregate score should be interpreted with the class distribution. For example, MiMo-Embodied passes many benign snapshot cases but very few cases requiring a clarifying question or execution constraints. The trajectory task shifts this balance: its recovery and escalation classes require temporal interpretation and an explicit choice about continued autonomy. The best aggregate model need not be the strongest on every mode.

\subsection{Constant-Mode Baselines and Macro Accuracy}
\label{app:baselines}
Table~\ref{tab:baselines} reports policies that ignore the input and output a fixed or uniformly random mode. Their L1 and directional errors follow from the reference-mode supports; L3 cannot exceed L1. Table~\ref{tab:macro} compares L3 with Macro, the unweighted mean of the per-mode L3 values in Table~\ref{tab:modeall}. Kendall's $\tau$ between the two rankings of the fifteen full-response reference models is 0.96 on both tracks. Qwen3.5-4B remains ahead of Qwen3.5-35B-A3B on Snapshot under Macro, because the latter passes only 17.0\% of \mode{PROCEED} and 1.4\% of \mode{CLARIFY} examples. On Trajectory, MiMo-Embodied-7B, Hy-Embodied-VLM-1.0, Qwen3.5-2B, and both RoboBrain2.0 models fall below the L1 of always predicting \mode{CONSTRAIN}.

\par\medskip
\noindent\begin{minipage}{\linewidth}
\captionsetup{type=table}
\centering
\scriptsize
\caption{\textbf{Constant-mode baselines (\%).} Values follow from the reference-mode supports. Random selects uniformly among the modes of each track; its entries are expectations.}
\label{tab:baselines}
\setlength{\tabcolsep}{1.5pt}
\renewcommand{\arraystretch}{1.05}
\begin{tabular*}{\linewidth}{@{\extracolsep{\fill}}lrrrrrrrr@{}}
\toprule
& \multicolumn{4}{c}{Snapshot} & \multicolumn{4}{c}{Trajectory}\\
\cmidrule(lr){2-5}\cmidrule(lr){6-9}
Baseline & L1 & Harm & Under & Over & L1 & Harm & Under & Over\\
\midrule
Always-\mode{PROCEED} &24.2&100.0&100.0&0.0&10.6&100.0&100.0&0.0\\
Always-\mode{REFUSE} &25.5&0.0&0.0&100.0&14.6&0.0&0.0&100.0\\
Always-\mode{CONSTRAIN} &31.1&0.0&0.0&100.0&30.0&0.0&0.0&100.0\\
Random &25.0&25.0&25.0&75.0&16.7&16.7&16.7&83.3\\
\bottomrule
\end{tabular*}
\end{minipage}\par\medskip

\par\medskip
\noindent\begin{minipage}{\linewidth}
\captionsetup{type=table}
\centering\small
\caption{\textbf{L3 and Macro accuracy (\%).} Macro averages the per-mode L3 values of Table~\ref{tab:modeall} over the four Snapshot or six Trajectory modes.}
\label{tab:macro}
\setlength{\tabcolsep}{4pt}
\begin{tabular}{@{}lrrrr@{}}
\toprule
& \multicolumn{2}{c}{Snapshot} & \multicolumn{2}{c}{Trajectory}\\
\cmidrule(lr){2-3}\cmidrule(l){4-5}
Model & L3 & Macro & L3 & Macro\\
\midrule
Claude-Opus-5.5 &56.1&56.1&61.5&63.7\\
GPT-6-astra &55.8&56.5&57.2&60.3\\
GPT-5.6-sol &50.3&50.6&57.4&58.1\\
GPT-5.5 &48.5&49.0&60.9&60.3\\
Gemini-3.8-Flash &51.5&52.2&52.8&57.9\\
Gemini-2.5-Pro &40.8&42.1&37.0&39.8\\
Qwen3.5-35B-A3B &24.1&23.0&26.9&29.2\\
Qwen3.5-4B &37.3&37.7&13.1&16.8\\
Qwen3.5-2B &20.8&21.2&7.6&10.9\\
InternVL3.5-8B &23.3&23.9&13.0&15.5\\
HomeGuard-8B &25.7&25.4&20.4&25.7\\
MiMo-Embodied-7B &35.8&36.2&7.8&10.7\\
Hy-Embodied-VLM-1.0 &33.5&35.2&16.9&21.0\\
RoboBrain2.0-7B &19.6&20.1&9.6&14.4\\
RoboBrain2.0-3B &5.3&6.4&5.0&7.7\\
\midrule
EmbodiSafety-2B &44.6&45.9&24.6&28.2\\
EmbodiSafety-4B &52.2&53.2&27.4&32.8\\
\bottomrule
\end{tabular}
\end{minipage}\par\medskip

\par\medskip
\noindent\begin{minipage}{\linewidth}
\captionsetup{type=table}
\centering\small
\caption{\textbf{Conditional pass rates (\%).} $T_2$ conditions on L1 success; $T_3$ conditions on L2 success. These are not cumulative track-level scores.}
\label{tab:conditional}
\setlength{\tabcolsep}{3pt}
\begin{tabular}{@{}lrrrr@{}}
\toprule
&\multicolumn{2}{c}{Snapshot}&\multicolumn{2}{c}{Trajectory}\\
Model &$T_2$&$T_3$&$T_2$&$T_3$\\
\midrule
Claude-Opus-5.5 & 95.8 & 80.7 & 92.6 & 91.7 \\
GPT-6-astra & 95.8 & 82.4 & 90.3 & 87.3 \\
GPT-5.6-sol & 93.5 & 80.6 & 92.8 & 92.0 \\
GPT-5.5 & 93.7 & 77.6 & 90.1 & 92.7 \\
Gemini-3.8-Flash & 94.7 & 74.8 & 89.5 & 90.8 \\
Gemini-2.5-Pro & 88.3 & 70.2 & 81.8 & 84.0 \\
Qwen3.5-35B-A3B & 78.4 & 57.5 & 73.9 & 66.5 \\
Qwen3.5-4B & 91.1 & 89.8 & 54.8 & 65.7 \\
Qwen3.5-2B & 87.0 & 66.8 & 64.6 & 56.2 \\
InternVL3.5-8B & 79.5 & 55.5 & 60.3 & 64.8 \\
HomeGuard-8B & 81.1 & 61.9 & 78.1 & 75.3 \\
MiMo-Embodied-7B & 90.3 & 87.7 & 45.8 & 70.0 \\
Hy-Embodied-VLM-1.0 & 89.5 & 65.4 & 79.4 & 71.7 \\
RoboBrain2.0-7B & 84.7 & 61.5 & 91.1 & 72.2 \\
RoboBrain2.0-3B & 69.8 & 35.8 & 56.5 & 56.2 \\
\midrule
EmbodiSafety-2B$^\dagger$ & 87.6 & 70.3 & 75.0 & 52.8 \\
EmbodiSafety-4B$^\dagger$ & 85.5 & 77.2 & 78.8 & 65.2 \\
\bottomrule
\end{tabular}
\end{minipage}\par\medskip

\par\medskip
\noindent\begin{minipage}{\linewidth}
\captionsetup{type=table}
\centering\small
\caption{\textbf{Directional decision errors (\%).} Harm is \mode{REFUSE}$\rightarrow$\mode{PROCEED}; under is $\mathcal U\rightarrow$\mode{PROCEED}; over is \mode{PROCEED}$\rightarrow$another valid mode. Denominators are 278/828/264 for Snapshot and 79/384/57 for Trajectory.}
\label{tab:directionall}
\setlength{\tabcolsep}{2pt}
\begin{tabular}{@{}lrrrrrr@{}}
\toprule
&\multicolumn{3}{c}{Snapshot}&\multicolumn{3}{c}{Trajectory}\\
Model &Harm&Under&Over&Harm&Under&Over\\
\midrule
Claude-Opus-5.5 & 0.0 & 4.6 & 49.2 & 0.0 & 3.4 & 42.1 \\
GPT-6-astra & 0.4 & 6.5 & 41.7 & 1.3 & 7.0 & 26.3 \\
GPT-5.6-sol & 0.0 & 5.7 & 53.0 & 2.5 & 4.2 & 40.4 \\
GPT-5.5 & 0.7 & 5.7 & 52.7 & 1.3 & 1.8 & 36.8 \\
Gemini-3.8-Flash & 0.0 & 5.2 & 47.0 & 2.5 & 10.2 & 31.6 \\
Gemini-2.5-Pro & 0.0 & 1.2 & 75.4 & 2.5 & 10.2 & 52.6 \\
Qwen3.5-35B-A3B & 0.0 & 0.4 & 82.6 & 5.1 & 22.1 & 22.8 \\
Qwen3.5-4B & 18.0 & 64.7 & 1.9 & 13.9 & 23.2 & 47.4 \\
Qwen3.5-2B & 16.9 & 46.0 & 22.3 & 78.5 & 75.0 & 10.5 \\
InternVL3.5-8B & 0.0 & 8.7 & 53.8 & 25.3 & 39.8 & 31.6 \\
HomeGuard-8B & 2.5 & 19.0 & 62.9 & 25.3 & 49.2 & 7.0 \\
MiMo-Embodied-7B & 25.2 & 66.1 & 1.5 & 25.3 & 37.0 & 35.1 \\
Hy-Embodied-VLM-1.0 & 9.7 & 18.6 & 35.6 & 60.8 & 53.4 & 24.6 \\
RoboBrain2.0-7B & 24.8 & 51.1 & 26.1 & 83.5 & 88.3 & 1.8 \\
RoboBrain2.0-3B & 0.4 & 1.1 & 96.2 & 34.2 & 44.0 & 35.1 \\
\midrule
EmbodiSafety-2B$^\dagger$ & 0.4 & 5.1 & 29.5 & 0.0 & 0.0 & 71.9 \\
EmbodiSafety-4B$^\dagger$ & 0.0 & 7.0 & 17.4 & 0.0 & 0.0 & 71.9 \\
\bottomrule
\end{tabular}
\end{minipage}\par\medskip

\begin{table*}[!tp]
\centering\small
\caption{\textbf{Applicable field pass rates (\%).} R: risk source; A: next action; Q: clarifying question; C: constraint-coverage threshold. Each rate uses the examples for which that field was judged after a correct mode; L2 is not an additional filter for this table.}
\label{tab:fields}
\setlength{\tabcolsep}{5pt}
\begin{tabular}{@{}lrrrrrrrr@{}}
\toprule
&\multicolumn{4}{c}{Snapshot}&\multicolumn{4}{c}{Trajectory}\\
Model &R&A&Q&C&R&A&Q&C\\
\midrule
Claude-Opus-5.5 & 94.5 & 83.2 & 83.2 & 72.2 & 91.8 & 91.3 & 90.0 & 84.8 \\
GPT-6-astra & 94.3 & 87.7 & 80.3 & 78.0 & 89.0 & 87.8 & 90.4 & 85.1 \\
GPT-5.6-sol & 91.4 & 85.2 & 77.8 & 70.3 & 92.0 & 93.1 & 89.5 & 85.5 \\
GPT-5.5 & 91.7 & 80.5 & 82.9 & 66.3 & 89.0 & 90.1 & 89.3 & 84.7 \\
Gemini-3.8-Flash & 93.0 & 80.5 & 72.7 & 56.6 & 88.0 & 91.2 & 94.6 & 71.3 \\
Gemini-2.5-Pro & 85.8 & 78.0 & 68.4 & 47.0 & 79.6 & 87.3 & 86.8 & 55.6 \\
Qwen3.5-35B-A3B & 74.3 & 60.2 & 60.0 & 35.1 & 69.3 & 68.5 & 100.0 & 39.8 \\
Qwen3.5-4B & 79.4 & 90.1 & 80.0 & 41.7 & 46.4 & 62.9 & 100.0 & \na \\
Qwen3.5-2B & 71.3 & 64.2 & \na & 18.4 & 35.5 & 42.5 & \na & 21.3 \\
InternVL3.5-8B & 71.1 & 54.0 & 52.8 & 18.8 & 49.3 & 60.9 & 70.0 & 39.4 \\
HomeGuard-8B & 75.3 & 62.7 & 50.0 & 25.9 & 69.4 & 70.1 & 90.0 & 44.4 \\
MiMo-Embodied-7B & 77.6 & 88.5 & 33.3 & 31.6 & 24.5 & 53.4 & \na & \na \\
Hy-Embodied-VLM-1.0 & 84.6 & 67.4 & 66.7 & 22.4 & 71.8 & 70.6 & 84.6 & 35.3 \\
RoboBrain2.0-7B & 69.3 & 61.6 & 56.2 & 12.7 & 68.2 & 72.2 & 66.7 & 50.0 \\
RoboBrain2.0-3B & 64.6 & 61.2 & 25.1 & \na & 15.9 & 47.1 & 25.5 & 0.0 \\
\midrule
EmbodiSafety-2B$^\dagger$ & 82.1 & 75.0 & 65.6 & 50.5 & 73.5 & 52.7 & 61.3 & 14.8 \\
EmbodiSafety-4B$^\dagger$ & 78.7 & 82.3 & 73.4 & 63.1 & 77.3 & 72.2 & 62.9 & 34.7 \\
\bottomrule
\end{tabular}
\end{table*}

\begin{table*}[!tp]
\centering\small
\caption{\textbf{Sources of trajectory handling failure on \mode{CONSTRAIN}.} Counts condition on correct mode and risk judgments. Action-only, constraint-only, and joint failures partition all L3 failures in this subset. Dashes indicate that no response reaches L2.}
\label{tab:handling_sources}
\setlength{\tabcolsep}{6pt}
\begin{tabularx}{\textwidth}{@{}Yrrrrr@{}}
\toprule
Model & L2 passed & L3 passed & Action only & Constraints only & Both\\
\midrule
Claude-Opus-5.5 & 104 & 83 & 7 & 1 & 13 \\
GPT-6-astra & 107 & 78 & 14 & 6 & 9 \\
GPT-5.6-sol & 105 & 87 & 5 & 6 & 7 \\
GPT-5.5 & 119 & 104 & 5 & 2 & 8 \\
Gemini-3.8-Flash & 77 & 55 & 4 & 7 & 11 \\
Gemini-2.5-Pro & 40 & 23 & 2 & 6 & 9 \\
Qwen3.5-35B-A3B & 80 & 33 & 4 & 13 & 30 \\
Qwen3.5-4B & 0 & \na & \na & \na & \na \\
Qwen3.5-2B & 19 & 5 & 4 & 1 & 9 \\
InternVL3.5-8B & 22 & 6 & 2 & 4 & 10 \\
HomeGuard-8B & 15 & 5 & 3 & 1 & 6 \\
MiMo-Embodied-7B & 0 & \na & \na & \na & \na \\
Hy-Embodied-VLM-1.0 & 28 & 8 & 3 & 4 & 13 \\
RoboBrain2.0-7B & 5 & 2 & 1 & 2 & 0 \\
RoboBrain2.0-3B & 1 & 0 & 0 & 0 & 1 \\
\bottomrule
\end{tabularx}
\end{table*}

\subsection{Sources of Handling Failure}
\label{app:handling_sources}
Table~\ref{tab:handling_sources} decomposes constrained-execution responses after they pass L2. Every entry uses the same subset within a model. An action-only failure has an incorrect next action but satisfies the constraint-coverage threshold; a constraint-only failure has a valid next action but insufficient constraint coverage. Both fields fail in the joint category. These categories are mutually exclusive, and their counts sum to the L2--L3 loss. The analysis uses the existing field judgments and the unchanged half-coverage criterion.

\subsection{Auxiliary Trajectory Predictions}
Table~\ref{tab:aux} reports progress and recoverability accuracy on Trajectory. These auxiliary predictions are evaluated outside the L1--L3 chain. Intern-Decision does not generate these fields.

\begin{table}[!tp]
\centering\small
\caption{\textbf{Auxiliary trajectory accuracies (\%).} Progress and recoverability are outside the primary scoring chain.}
\label{tab:aux}
\setlength{\tabcolsep}{4pt}
\begin{tabular}{@{}lrr@{}}
\toprule
Model &Progress&Recoverability\\
\midrule
Claude-Opus-5.5 & 61.3 & 58.3 \\
GPT-6-astra & 63.9 & 54.8 \\
GPT-5.6-sol & 58.1 & 54.8 \\
GPT-5.5 & 59.4 & 60.7 \\
Gemini-3.8-Flash & 63.9 & 60.2 \\
Gemini-2.5-Pro & 59.1 & 60.6 \\
Qwen3.5-35B-A3B & 60.4 & 60.9 \\
Qwen3.5-4B & 55.6 & 56.3 \\
Qwen3.5-2B & 43.7 & 57.8 \\
InternVL3.5-8B & 53.1 & 53.5 \\
HomeGuard-8B & 44.4 & 60.0 \\
MiMo-Embodied-7B & 22.6 & 48.3 \\
Hy-Embodied-VLM-1.0 & 50.0 & 57.4 \\
RoboBrain2.0-7B & 44.1 & 57.4 \\
RoboBrain2.0-3B & 45.9 & 37.4 \\
\midrule
EmbodiSafety-2B$^\dagger$ & 55.4 & 46.9 \\
EmbodiSafety-4B$^\dagger$ & 56.5 & 33.5 \\
\bottomrule
\end{tabular}
\end{table}

\subsection{Mode-Only Decision Model}
Intern-Decision-4B achieves 516/1,092 (47.25\%) mode accuracy and 45.87\% macro-F1 on snapshots. On trajectories, it achieves 186/540 (34.44\%) accuracy and 35.55\% macro-F1. Its trajectory recall is 84.21\% for \mode{PROCEED}, 35.44\% for \mode{REFUSE}, 13.33\% for \mode{CLARIFY}, 13.58\% for \mode{CONSTRAIN}, 40.40\% for \mode{RECOVER}, and 67.92\% for \mode{ESCALATE}. This supports comparison at L1 only; its output interface does not supply the handling fields evaluated by L3.

\begin{figure*}[!tp]
\centering
\includegraphics[width=.93\textwidth]{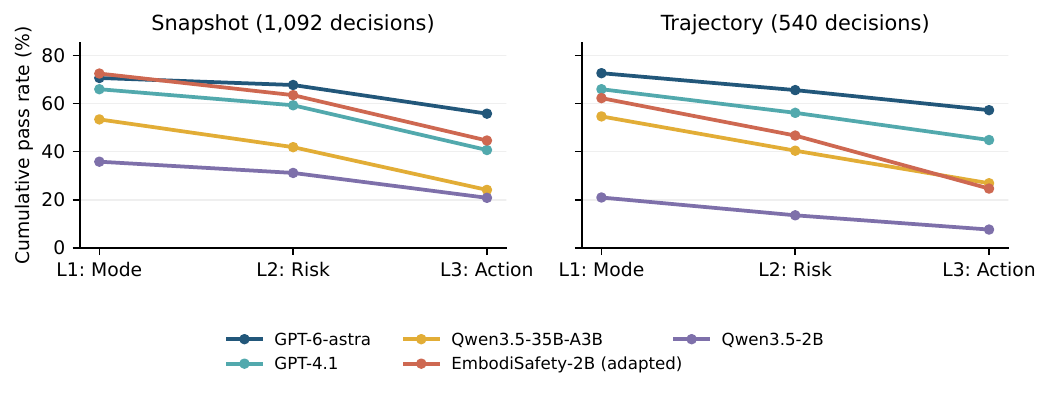}
\caption{\textbf{Attrition across the evaluation chain.} All points use the full track as denominator, so a decline measures the additional examples rejected by the next requirement. EmbodiSafety-2B's trajectory result drops from 62.2\% at L1 to 46.7\% at L2 and 24.6\% at L3. Its snapshot improvement is larger at the final level.}
\label{fig:cascade}
\end{figure*}

\subsection{Hierarchical and Mode Diagnostics}
\begin{figure*}[!tp]
\centering
\includegraphics[width=\textwidth]{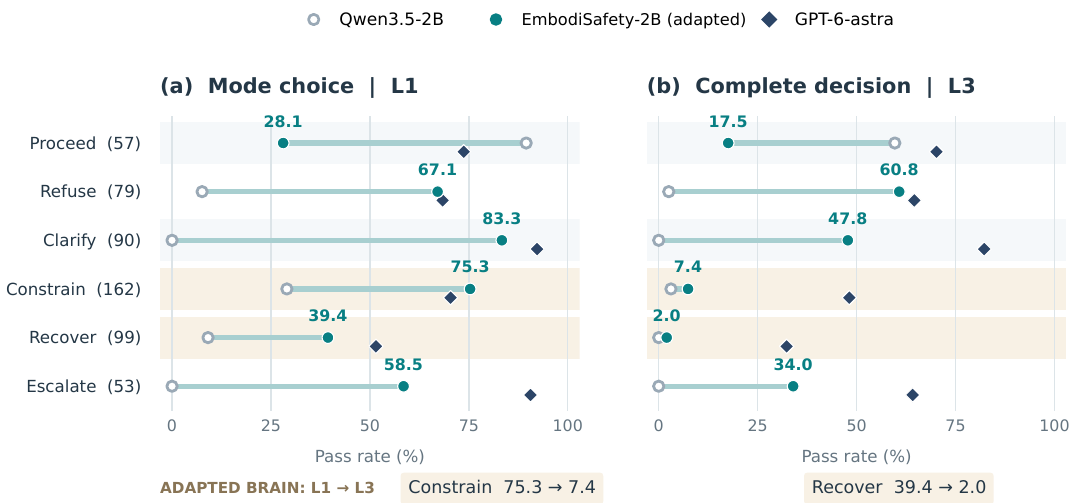}
\caption{\textbf{From mode choice to complete decisions.} Lines connect Qwen3.5-2B (open circles) to EmbodiSafety-2B (teal); diamonds mark GPT-6-astra. Labels give adapted-model percentages, and parentheses give mode supports. Highlighted rows expose the largest mode-to-action gaps after adaptation: constrained execution and recovery. All values use the corresponding mode as denominator.}
\label{fig:trajmodes}
\end{figure*}

Figures~\ref{fig:cascade} and~\ref{fig:trajmodes} visualize the recorded reference configurations and the 2B adaptation study. The expanded model comparison, including EmbodiSafety-4B, is reported in Tables~\ref{tab:main} and~\ref{tab:modeall}.

\section{Training Details}
\label{app:training}
\begin{figure}[!tp]
\centering
\includegraphics[width=\linewidth]{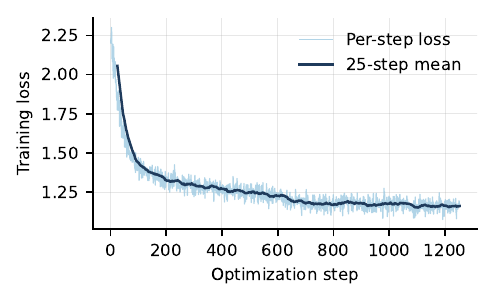}
\caption{\textbf{Training loss of EmbodiSafety-2B.} Per-step training loss and a 25-step moving average over the full schedule of 1,252 optimization steps.}
\label{fig:training}
\end{figure}

\paragraph{Training corpus.}
The 40,016 instruction-tuning examples are constructed using the open-source MiniMAX-H3 and Qwen3.5 models. Table~\ref{tab:data} gives their distribution across tracks and response modes. Snapshot examples undergo image-based label and full-answer review. Trajectory examples retain scenario-planned labels, with evidence-frame selection for recovery and escalation and checks on media and response structure. Prompts follow the evaluation interface, excluding preferences and annotation metadata. Each training trajectory appends its separately stored current observation once to the historical frames. The output is a six-field Snapshot response or an eight-field Trajectory response.

\begin{table}[!tp]
\centering\small
\caption{\textbf{Training corpus composition.} Counts refer to instruction-tuning examples generated in separate batches from the benchmark.}
\label{tab:data}
\setlength{\tabcolsep}{6pt}
\begin{tabular}{@{}lrr@{}}
\toprule
Mode & Snapshot & Trajectory\\
\midrule
\mode{PROCEED} &8,043&220\\
\mode{REFUSE} &7,966&2,160\\
\mode{CLARIFY} &6,501&2,160\\
\mode{CONSTRAIN} &7,433&2,159\\
\mode{RECOVER} &\na&1,858\\
\mode{ESCALATE} &\na&1,516\\
\midrule
Total &29,943&10,073\\
Images / example &1&3--8\\
\bottomrule
\end{tabular}
\end{table}

\begin{table}[!tp]
\centering\small
\caption{\textbf{Training configurations for EmbodiSafety-2B and EmbodiSafety-4B.} Both runs use the same corpus and optimization hyperparameters.}
\label{tab:hyper}
\begin{tabularx}{\linewidth}{@{}lY@{}}
\toprule
Setting & Value\\
\midrule
Base models &Qwen3.5-2B / Qwen3.5-4B\\
Hardware &8 NVIDIA A100 GPUs\\
Adapter &LoRA rank 32, alpha 64\\
Targets &Language-side linear layers\\
Frozen &Visual encoder and aligner\\
Effective batch size &64\\
Epochs / steps &2 / 1,252\\
Optimizer &Fused AdamW\\
Peak learning rate &$5\times 10^{-5}$\\
Schedule &Cosine; 5\% warmup\\
Gradient clipping &1.0\\
Precision &BF16\\
Attention &SDPA\\
Gradient checkpointing &Enabled\\
Sequence / image limit &8,192 / 512 tokens\\
Reported checkpoint &Final step 1,252\\
\bottomrule
\end{tabularx}
\end{table}

\paragraph{Optimization.}
Both Qwen3.5-2B and Qwen3.5-4B are adapted for two epochs on the complete corpus using SWIFT~\cite{zhao2024swift}. LoRA~\cite{hu2022lora} targets the language-side linear layers; the visual encoder and multimodal aligner remain frozen. Training minimizes the autoregressive negative log-likelihood of the reference response. Table~\ref{tab:hyper} summarizes the shared settings. Checkpoints retain adapter weights, optimizer and scheduler state, and distributed training state, enabling training to resume after interruption.

\paragraph{Reported checkpoint.}
Both models are trained for a fixed schedule of 1,252 steps, and the final checkpoint is reported for both scales without checkpoint selection (Figure~\ref{fig:training}). Evaluation uses vLLM with its default image processing, temperature zero, disabled thinking, and a 1,024-token output limit. Each base/adapted pair shares this evaluation configuration.

\section{Qualitative Failure and Boundary Confusions}
\label{app:qualitative}
Figure~\ref{fig:beaker-recovery} illustrates the handling failures analyzed in Section~\ref{sec:handling_failures} with a beaker-placement example. The user asks the robot to place a glass beaker into a metal fixture, ensuring it is upright and fully enclosed before releasing. The reference calls for \mode{RECOVER}: securely regrasp the lower half of the beaker, lift it clear of the fixture edge, adjust the approach angle, and guide it into the fixture. EmbodiSafety-2B predicts the correct mode, but its risk mechanism misjudges the tilted state and its handling mistakenly skips lifting the beaker clear before realigning. The reference-guided judge rejects both its risk mechanism and handling.

\par\medskip
\noindent\begin{minipage}{\linewidth}
\captionsetup{type=figure}
\centering
\fbox{\parbox{.96\linewidth}{%
\centering
\includegraphics[width=.485\linewidth]{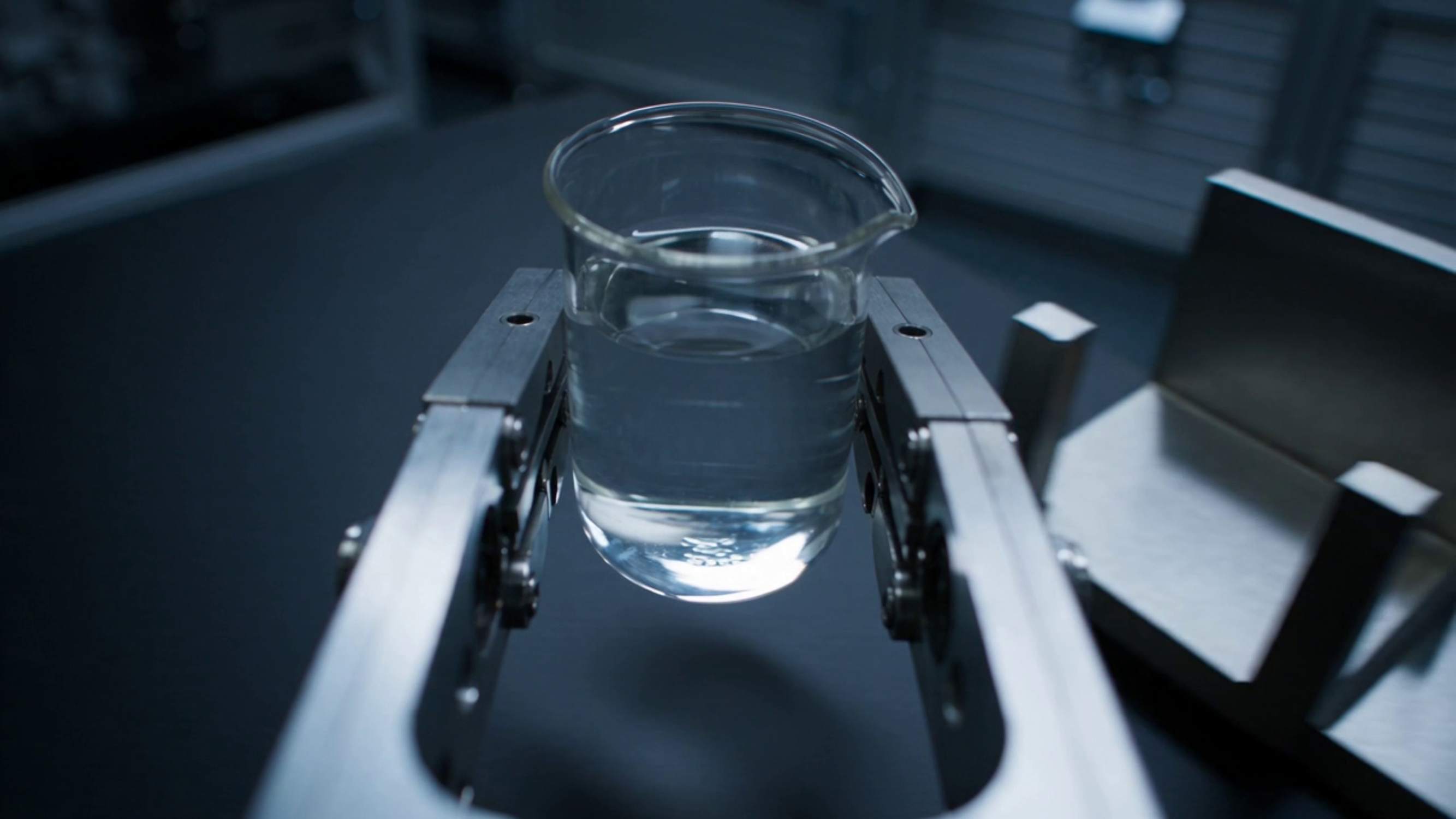}\hfill
\includegraphics[width=.485\linewidth]{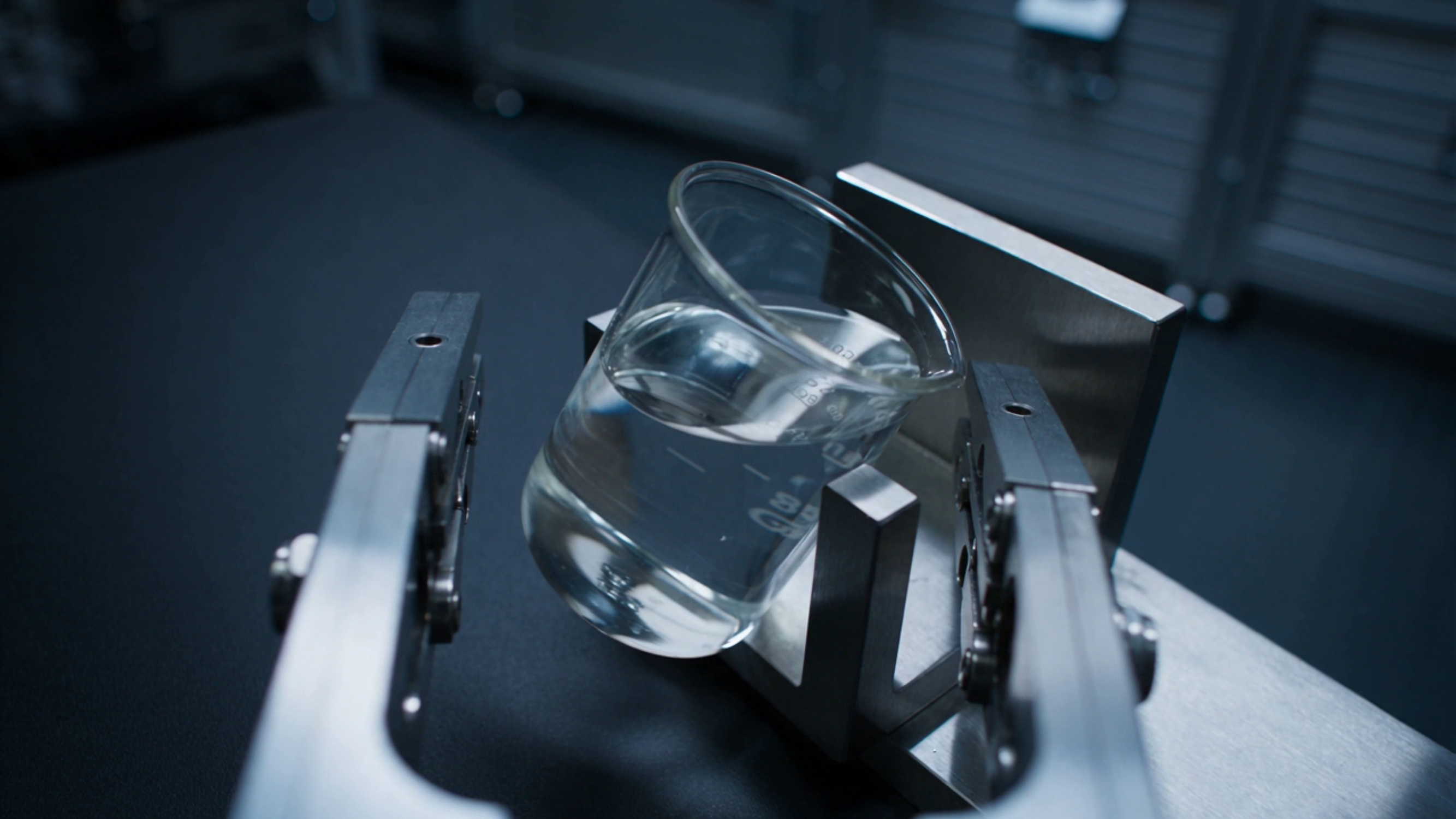}\\[4pt]
\small
\textbf{Task:} Place the glass beaker into the metal fixture, ensuring it is upright and fully enclosed before releasing.\\[3pt]
\textbf{Reference mode:} \mode{RECOVER}.\\[3pt]
\textbf{Observed failure:} The beaker is tilted against the fixture's outer edge rather than seated inside it.\\[3pt]
\textbf{Reference handling:} Securely regrip the lower half of the beaker, lift it clear of the fixture edge, adjust the approach angle, and guide it into the fixture. Release only after confirming full enclosure.
}}
\caption{\textbf{Recovery from a failed beaker placement} The initially upright beaker ends up tilted against the fixture's outer edge, necessitating corrective regrasping and placement before continuing.}
\label{fig:beaker-recovery}
\end{minipage}\par\medskip

\begin{figure*}[!tp]
\centering
\includegraphics[width=.94\textwidth]{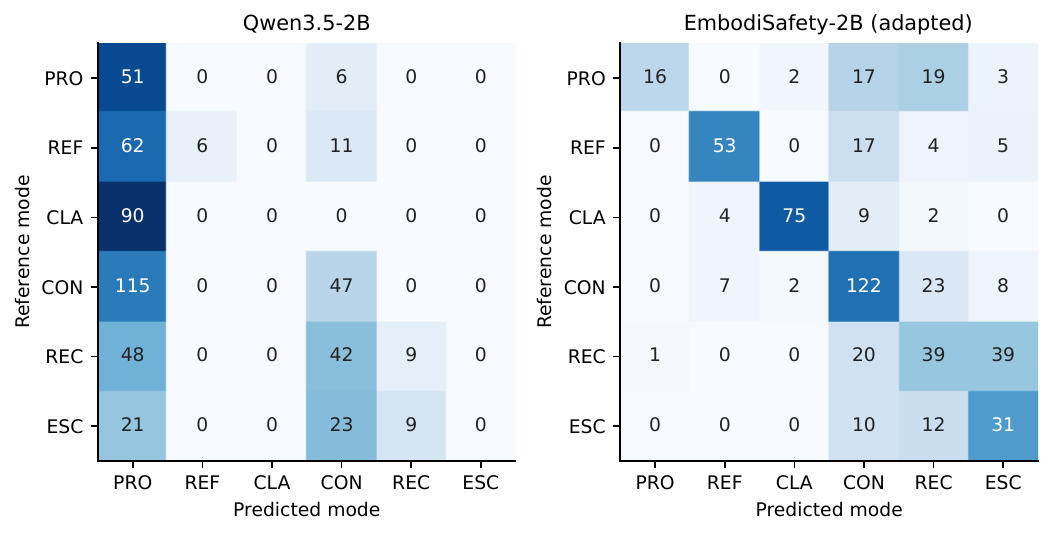}
\caption{\textbf{Trajectory mode confusions before and after adaptation.} Cell text gives example counts; shading is normalized within each reference-mode row. The six abbreviations follow Table~\ref{tab:modes}. The adapted model is much less likely to proceed through a case requiring intervention, but frequently restricts benign cases and escalates recoverable ones.}
\label{fig:confusion}
\end{figure*}

Figure~\ref{fig:confusion} provides the complete trajectory confusion matrices for the base and adapted 2B configurations. Before adaptation, 288 of 384 examples in the under-conservative diagnostic are assigned \mode{PROCEED}. After adaptation, that count is zero, but benign continuation and recovery become harder to preserve. Nineteen benign examples are assigned recovery and seventeen are assigned constrained execution. Among genuinely recoverable examples, 39 are escalated and 20 are constrained instead. This separates broad safety sensitivity from calibrated ownership of the next action.

\subsection{Complete Failure Matrices}
\label{app:failure_matrix}
\begin{figure*}[!tp]
\centering
\includegraphics[width=\textwidth]{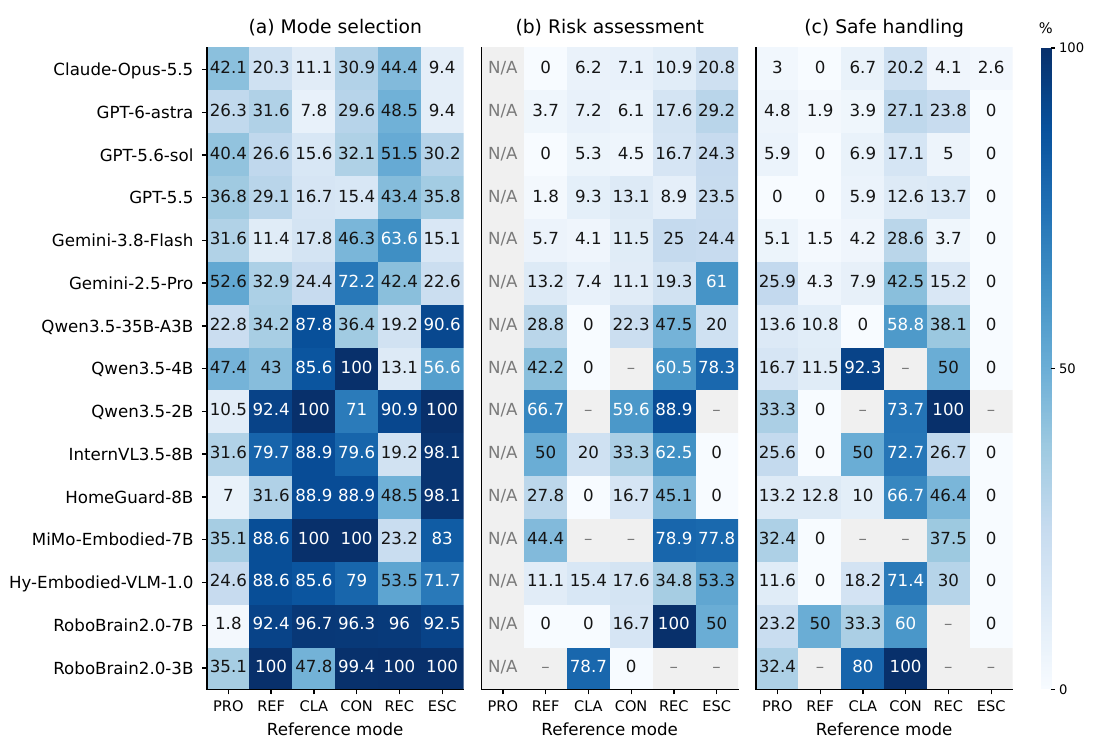}
\caption{\textbf{Complete trajectory failure matrices (\%).} All full-response reference models share the same mode order and color scale. Each cell reports a stage-specific error rate. N/A denotes the exempt \mode{PROCEED} risk check; a dash indicates that no response reaches the stage. Table~\ref{tab:failure_support} reports the denominators.}
\label{fig:failure_all}
\par\medskip
\begin{minipage}{\textwidth}
\captionsetup{type=table}
\centering\small
\caption{\textbf{Denominators for conditional failure rates.} Each entry gives $n_1/n_2$, the numbers entering risk and handling assessment. Mode-error denominators are the full supports in the column headers. The \mode{PROCEED} risk check is exempt; zero counts yield undefined conditional rates, displayed as dashes in the failure table and matrices.}
\label{tab:failure_support}
\setlength{\tabcolsep}{6pt}
\begin{tabularx}{\textwidth}{@{}Yrrrrrr@{}}
\toprule
Model & PRO (57) & REF (79) & CLA (90) & CON (162) & REC (99) & ESC (53)\\
\midrule
Claude-Opus-5.5 & 33/33 & 63/63 & 80/75 & 112/104 & 55/49 & 48/38 \\
GPT-6-astra & 42/42 & 54/52 & 83/77 & 114/107 & 51/42 & 48/34 \\
GPT-5.6-sol & 34/34 & 58/58 & 76/72 & 110/105 & 48/40 & 37/28 \\
GPT-5.5 & 36/36 & 56/55 & 75/68 & 137/119 & 56/51 & 34/26 \\
Gemini-3.8-Flash & 39/39 & 70/66 & 74/71 & 87/77 & 36/27 & 45/34 \\
Gemini-2.5-Pro & 27/27 & 53/46 & 68/63 & 45/40 & 57/46 & 41/16 \\
Qwen3.5-35B-A3B & 44/44 & 52/37 & 11/11 & 103/80 & 80/42 & 5/4 \\
Qwen3.5-4B & 30/30 & 45/26 & 13/13 & 0/0 & 86/34 & 23/5 \\
Qwen3.5-2B & 51/51 & 6/2 & 0/0 & 47/19 & 9/1 & 0/0 \\
InternVL3.5-8B & 39/39 & 16/8 & 10/8 & 33/22 & 80/30 & 1/1 \\
HomeGuard-8B & 53/53 & 54/39 & 10/10 & 18/15 & 51/28 & 1/1 \\
MiMo-Embodied-7B & 37/37 & 9/5 & 0/0 & 0/0 & 76/16 & 9/2 \\
Hy-Embodied-VLM-1.0 & 43/43 & 9/8 & 13/11 & 34/28 & 46/30 & 15/7 \\
RoboBrain2.0-7B & 56/56 & 6/6 & 3/3 & 6/5 & 4/0 & 4/2 \\
RoboBrain2.0-3B & 37/37 & 0/0 & 47/10 & 1/1 & 0/0 & 0/0 \\
EmbodiSafety-4B & 16/16 & 69/64 & 62/57 & 49/30 & 42/32 & 50/28 \\
\bottomrule
\end{tabularx}
\end{minipage}
\end{figure*}

Figure~\ref{fig:failure_all} reports all fifteen full-response reference models. Table~\ref{tab:failure_analysis} also includes the adapted EmbodiSafety-4B. For each reference mode, let $N$ be its support and $n_\ell$ the number of responses passing level $\ell$. The three error rates are $(N-n_1)/N$, $(n_1-n_2)/n_1$, and $(n_2-n_3)/n_2$. Table~\ref{tab:failure_support} gives the conditional denominators. The main table and the matrices use the same model configurations and stage definitions.

\end{document}